\documentclass[10pt,journal,compsoc]{IEEEtran}

\usepackage[utf8]{inputenc}
\usepackage[T1]{fontenc}
\usepackage{url}
\usepackage{booktabs}
\usepackage{amsfonts}
\usepackage{amsmath}
\usepackage{nicefrac}
\usepackage{microtype}
\usepackage{amsthm}
\usepackage{amssymb}
\usepackage{rotating}
\usepackage{graphicx}
\usepackage{caption}
\usepackage{wrapfig}
\usepackage{bm}
\usepackage{xcolor}            % xcolor 已包含 color 的所有功能
\usepackage{colortbl}
\usepackage{multirow}
\usepackage{mathtools}
\usepackage{lscape}
\usepackage{hyperref}
\usepackage{bbm}

\newcommand{\eat}[1]{}

\usepackage{tikz}
\usepackage{svg}
\usepackage{bbding}
\usepackage{cleveref}
\usepackage{algorithmic}
\usepackage{algorithm}
\usepackage{makecell}
\usepackage{setspace}
\usepackage{soul}              % 如果 preamble 里保留了 \st

\usepackage[caption=false,font=normalsize,labelfont=sf,textfont=sf]{subfig}

\usepackage{textcomp}
\usepackage{stfloats}
\usepackage{verbatim}
\usepackage[nocompress]{cite}   % compsoc 要求 nocompress
\usepackage{stackengine}
\usepackage{diagbox}
\usepackage{pifont}

\definecolor{citeblue}{HTML}{0071bc}
\hypersetup{
  colorlinks,
  linkcolor = citeblue,
  citecolor = citeblue,
}

\definecolor{darkgreen}{rgb}{0.0, 0.5, 0.0}
\definecolor{darkred}{rgb}{0.55, 0.0, 0.0}

\definecolor{myred}{rgb}{0.8,0,0}
\definecolor{mygreen}{rgb}{0,0.8,0}

\definecolor{red}{rgb}{0.95,0.84,0}
\definecolor{orange}{rgb}{0.82,0.82,0.82}
\definecolor{yellow}{rgb}{0.78,0.62,0.48}

\definecolor{junglegreen}{rgb}{0.02, 1.0, 0.27}

\definecolor{mypurple}{rgb}{ 0.174, 0.190, 0.876}

\newcommand{\ffst}[1]{\cellcolor{red}#1}
\newcommand{\fsnd}[1]{\cellcolor{orange}#1}
\newcommand{\ftrd}[1]{\cellcolor{yellow}#1}

\newcommand{\fst}[1]{\colorbox{red}{#1}}
\newcommand{\snd}[1]{\colorbox{orange}{#1}}
\newcommand{\trd}[1]{\colorbox{yellow}{#1}}

\begin{document}

\title{MUA: Mobile Ultra-detailed Animatable Avatars}

\author{Heming Zhu, Guoxing Sun, Marc Habermann$^{\dag}$
\IEEEcompsocitemizethanks{
  \IEEEcompsocthanksitem H. Zhu, G. Sun and M. Habermann are with the Max Planck Institute for Informatics, Saarland Informatics Campus, Germany. (email: \{hezhu, gsun, mhaberma\}@mpi-inf.mpg.de).
  \IEEEcompsocthanksitem $\dag$ denotes corresponding author.
}}

\markboth{}{}

\IEEEtitleabstractindextext{%
%
%%%%%%%%%%%%%%%%%%%%%%
%
\begin{abstract}
Building photorealistic, animatable full-body digital humans remains a longstanding challenge in computer graphics and vision. 
Recent advances in animatable avatar modeling have largely progressed along two directions: improving the fidelity of dynamic geometry and appearance, or reducing computational complexity to enable deployment on resource-constrained platforms, e.g., VR headsets. 
However, existing approaches fail to achieve both goals simultaneously:
Ultra-high-fidelity avatars typically require substantial computation on server-class GPUs, whereas lightweight avatars often suffer from limited surface dynamics, reduced appearance details, and noticeable artifacts.
To bridge this gap, we propose a novel animatable avatar representation, termed Wavelet-guided Multi-level Spatial Factorized Blendshapes, and a corresponding distillation pipeline that transfers motion-aware clothing dynamics and fine-grained appearance details from a pre-trained ultra-high-quality avatar model into a compact, efficient representation.
By coupling multi-level wavelet spectral decomposition with low-rank structural factorization in texture space, our method achieves up to 2000× lower computational cost and a 10× smaller model size than the original high-quality teacher avatar model, while preserving visually plausible dynamics and appearance details closely resemble those of the teacher model.
Extensive comparisons with state-of-the-art methods show that our approach significantly outperforms existing avatar approaches designed for mobile settings and achieves comparable or superior rendering quality to most approaches that can only run on servers.
Importantly, our representation substantially improves the practicality of high-fidelity avatars for immersive applications, achieving over 180 FPS on a desktop PC and real-time native on-device performance at 24 FPS on a standalone Meta Quest 3.
Additional results and resources are available on the project page: \href{https://vcai.mpi-inf.mpg.de/projects/MUA/}{https://vcai.mpi-inf.mpg.de/projects/MUA}.
\end{abstract}

}

\maketitle
\IEEEdisplaynontitleabstractindextext
\IEEEpeerreviewmaketitle

%
%%%%%%%%%%%%%%%%%%%%%%%%%%%%%%%%%%%%%%%%%%%%%%%%%%%%%%%%%%%%%%%%%%%%%%%
%
\section{Introduction} \label{sec:intro}
\IEEEPARstart{C}{r}eating photorealistic clothed humans that can be animated directly from skeletal poses remains a longstanding challenge in computer graphics and vision.
One of the primary usage scenarios is virtual and mixed reality, where vivid and immersive experiences through VR headsets can enhance decision-making~\cite{chen2017exploring} and facilitate education~\cite{sayffaerth2025educational}.
\par
Driven by recent advances in neural 3D representations~\cite{mildenhall2020nerf, kerbl20233d, neus2}, 
the fidelity of animatable clothed human avatars~\cite{li2024animatable,habermann2023hdhumans} has significantly improved.
They typically attach neural representations onto a drivable human skeleton~\cite{li2022tava}, a parametric human body model~\cite{ARAH}, or a deformable template mesh~\cite{Pang_2024_CVPR}.
However, due to the inaccuracy of surface tracking, these methods fall short in \textbf{capturing and synthesizing high-frequency details}, e.g., fine texture patterns and detailed geometry deformations.
To address this issue, UMA~\cite{zhu2025ultra} leverages guidance from video point trackers~\cite{karaev2024cotracker} to improve geometric alignment for both the coarse template mesh and the fine 3D Gaussian Splat geometry.
While UMA recovers realistic clothing dynamics and intricate wrinkle details, it requires a dedicated server equipped with high-end graphics cards and still merely meets the real-time requirement of $20$ FPS.
This restricts its deployment on more accessible platforms, such as personal computers with consumer-grade graphics cards, as well as even more lightweight devices such as VR and AR headsets.
\par 
To deploy the animatable avatars on devices with limited computational budgets, SqueezeMe~\cite{iandola2025squeezeme} proposes an efficient formulation of pose-dependent 3DGS correctives, whereas TaoAvatar~\cite{chen2025taoavatar} distills a high-capacity deformation model into a lightweight avatar representation with learned 3DGS blendshapes.
Although both methods achieve real-time performance on mobile devices, they remain limited in capturing \textbf{large-scale motion-dependent clothing dynamics} and \textbf{intricate wrinkle details} under such constrained budgets, thereby significantly compromising both visual realism and the sense of immersion.
In particular, large-scale non-rigid garment deformations, such as dress swaying, are often reduced to rigid skinning-based deformation.
Meanwhile, fine-scale wrinkles exhibit little motion-dependent variation and often appear nearly baked into the surface.
This raises a fundamental research question: How can we build \textbf{photorealistic animatable avatars} that faithfully capture both large-scale clothing dynamics and fine-grained, zoom-in levels of appearance details while supporting \textbf{real-time inference on mobile devices} such as VR headsets? 
\par
To bring high-fidelity animatable avatars to customer-friendly devices, we propose \textbf{MUA}, i.e., \textbf{M}obile \textbf{U}ltra-detailed \textbf{A}nimatable Avatars.
Since the major computational bottleneck of UMA stems from its dense 2D UNet-based image-to-image translation network for producing high-resolution (i.e., $768 \times 768$) Gaussian texture maps, we instead adopt a learnable blendshape-based formulation, which has been shown effective for modeling dynamic animatable characters~\cite{chen2025taoavatar,SMPL-X:2019}.
However, defining blendshape bases directly over such full-resolution Gaussian splat textures inevitably results in a prohibitively large memory footprint, making this formulation unsuitable for mobile devices.
\par
Recent advances in factorized neural shape representations~\cite{chen2022tensorf} address this issue by showing that high-dimensional spatial signals can be represented through the outer product of lower-dimensional components, thereby drastically reducing the number of parameters.
Yet, naively applying factorization to full-resolution Gaussian splat textures is fundamentally ill-suited to our setting, as their mixed-frequency structure violates the low-rank and sparsity assumptions for effective factorization.

To this end, we propose \textbf{wavelet-guided multi-resolution factorized blendshapes}, which decompose the full-resolution Gaussian splat texture into multi-level wavelet subbands and model each subband with a tailored representation that matches its structural characteristics.
Specifically, the highly compact low-frequency subband, e.g., $48\times48$, is modeled directly with 2D blendshapes, while the detail subbands are represented with factorized formulations due to their inherent spatial sparsity.
The predicted subbands are fused via the inverse discrete wavelet transform (IDWT) to reconstruct the full-resolution Gaussian splat texture, enabling expressive modeling of both large-scale motion-driven deformations and fine-grained details at low computational cost (less than one GFLOP).

To validate the effectiveness of our method, we compare it against both server and mobile-based approaches in terms of reconstruction and generation quality, computational cost, and parameter count.
Our method outperforms mobile-based approaches by a clear margin, while achieving performance comparable to, or even better than, server-based approaches, despite requiring three orders of magnitude less computation at test time.
Moreover, we conduct extensive ablation studies on several key design choices, including the representation domain, shape representation, and network architecture.
These ablations provide insights for future research on expressive and lightweight animatable avatar representations.
Finally, we build an interactive system based on the proposed model, achieving over $180$ FPS on a desktop workstation and real-time performance ($24$ FPS) on a Meta Quest 3 headset.

%
%%%%%%%%%%%%%%%%%%%%%%%%%%%%%%%%%%%%%%
%
\begin{figure*}[tp]
    \centering
    \includegraphics[width=\linewidth]{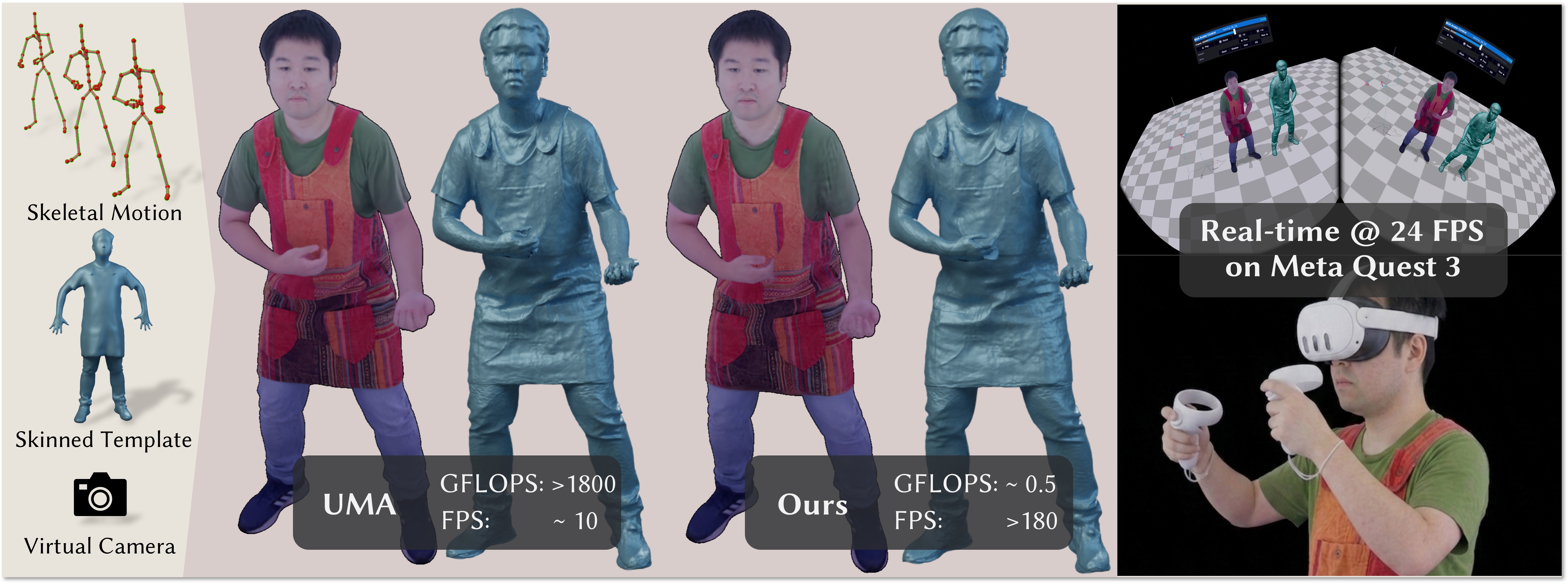}
    \vspace{-15pt}
    \caption{
    Given skeletal poses and a virtual camera as inputs, \textbf{MUA} produces photorealistic renderings and detailed geometry of animatable clothed humans.
    By distilling the ultra-high-quality teacher avatar model, i.e., UMA, into a compact student representation, \textbf{MUA} preserves large-scale clothing dynamics together with fine geometric and appearance details, while reducing computation by three orders of magnitude and achieving over 180 FPS on a personal computer.
    Moreover, \textbf{MUA} enables real-time on-device inference at $24$ FPS on a standalone Meta Quest 3 headset, advancing the practical deployment of highly detailed animatable avatars on VR headsets and other computation-constrained mobile platforms.
    }
    \label{fig:teaser}

\end{figure*}
%
%%%%%%%%%%%%%%%%%%%%%%%%%%%%%%%%%%%%%%
%
%
Our contributions are summarized as follows:
\begin{itemize}
\item We propose a novel compact representation, namely \textbf{wavelet-guided multi-resolution factorized blendshapes}, for modeling high-quality animatable avatars.
The proposed representation preserves the clothing dynamics and fine-grained appearance details of an ultra-high-quality teacher model, while reducing the computational cost by three orders of magnitude and the parameter count to one-tenth.
\item We introduce a distillation pipeline, i.e., \textbf{MUA}, that transfers motion-dependent geometric deformations and appearance details from a high-fidelity teacher model into the proposed compact student representation.
\item We develop a real-time VR system based on the proposed representation, achieving $24$ FPS with fully on-device inference on Meta Quest 3 headset and $90$ FPS in a PC-connected mode, demonstrating the practical applicability of our method.
\item We conduct extensive comparisons with the state-of-the-art animatable avatar methods, together with comprehensive ablation studies on key design choices, providing practical insights for future research on lightweight, high-quality animatable avatars.
\end{itemize}

%
%%%%%%%%%%%%%%%%%%%%%%%%%%%%%%%%%%%%%%%%%%%%%%%%%%%%%%%%%%%%%%%%%%%%%%%
%
\section{Related Work} \label{sec:rw}
Our work focuses on animatable full-body clothed human avatars that take only skeletal pose as input during inference.
Accordingly, we do not consider methods for 4D replay~\cite{peng2021neuralbody,Lombardi2021MVP,wang2020learning,weng_humannerf_2022_cvpr,isik2023humanrf,xu2024representing,jiang2025reperformer}, which focuses on rendering pre-recorded sequences.
Furthermore, we do not cover reconstruction~\cite{xiang2021modeling,alldieck18b,alldieck19,habermann2020deepcap,habermann2019livecap,xiu2023econ,zhang2024sifu,zheng2025gstar,zhu2022registering} or image-based free-view rendering~\cite{kwon2021neural, wang2021ibrnet,Remelli2022TexelAligned,shetty2023holoported,sun2024metacap,sun2025real,zubekhin2025giga}, as they require image inputs at test time.

\subsection{Photorealistic Full-Body Avatar}
According to the underlying shape representations, photorealistic full-body avatars can be categorized into mesh-based, implicit-based, and point-based approaches.

\noindent\textbf{Mesh-based approaches.} Inherited from traditional animation pipelines, textured meshes remain the dominant representation due to their compatibility with most rendering and animation software such as Blender~\cite{blender}.
To generate pose-dependent textures and clothing wrinkles, early approaches adopt physical simulation~\cite{stoll2010video, guan2012drape}, video database retrieval~\cite{xu2011video}, or interpolation within texture stacks~\cite{casas14, shysheya2019textured}.
More recently, Bagautdinov et al.~\cite{bagautdinov2021driving,xiang2021modeling,xiang2022dressing} leverage neural networks to learn motion-dependent textures from multi-view videos.
Deep Dynamic Characters~\cite{habermann2021} models surface deformations with embedded deformation graphs~\cite{embedded} and capture dynamic appearance through motion-aware textures.
MeshAvatar~\cite{chen2024meshavatar} parameterizes both geometry and surface materials via orthogonal front-back feature maps.
However, bounded by the resolution of the template mesh, mesh-based approaches often fail to capture fine-grained geometric and appearance details.

\noindent\textbf{Implicit-based approaches.} 
Implicit-based approaches improve the capacity of mesh-based methods by coupling neural radiance fields (i.e., NeRF)~\cite{mildenhall2020nerf,wang2021neus} with drivable shape proxies, such as virtual bones~\cite{li2022tava}, parametric body meshes~\cite{loper15,STAR:2020,SMPL-X:2019,TotalCapture2018}, or person-specific template meshes~\cite{habermann2020deepcap}.
More specifically, to capture motion-dependent clothing deformations and dynamic appearance, recent works~\cite{liu2021neural, peng2021animatable, xu2021hnerf, NNA, zheng2023avatarrex, kwon2023deliffas} learn pose-conditioned NeRFs in a shared canonical space.
However, due to the massive spatial sampling for volume rendering and large per-sample MLP evaluations, it usually takes seconds to render a frame on server-class GPUs.
TriHuman~\cite{zhu2023trihuman} achieves real-time rendering on server-class GPUs by adopting a deformable texture-space triplane~\cite{Chan2022} which offloads the per-sample MLP evaluations to per-frame feature map computation.
Nevertheless, generating pose-dependent feature maps still relies on heavy 2D convolutional networks, which remain computationally expensive and hinder deployment on consumer-grade GPUs and VR headsets.

\noindent\textbf{Point-based approaches.} Compared to mesh-based representations, point-based approaches offer greater flexibility for modeling fine geometric details.
Moreover, their explicit formulation enables efficient and fast rendering.
SCALE~\cite{ma2021scale}, POP~\cite{ma2021power}, and FITE~\cite{lin2022learning} model non-rigid clothing deformations using dense point clouds defined in the UV space or in orthographic projections of parametric body models~\cite{loper2015smpl}.

In recent years, 3D Gaussian Splatting~\cite{kerbl20233d} has become a popular choice for animatable clothed avatars due to its real-time, high-fidelity rendering.
GART~\cite{lei2023gart}, 3DGS-Avatar~\cite{qian20233dgs}, GauHuman~\cite{hu2023gauhuman}, and HUGS~\cite{kocabas2023hugs} learn canonical Gaussian Splat avatars from monocular videos and animate them via linear blend skinning.
However, the fine details are missing since the Gaussian attributes are predicted directly through MLPs.
Therefore, ASH~\cite{Pang_2024_CVPR}, GaussianAvatar~\cite{hu2023gaussianavatar}, and Animatable Gaussians~\cite{li2024animatable} learn animatable characters with motion-aware appearance by leveraging convolutional neural networks in the UV space or in orthographic projections.
By addressing drifting surface correspondences over frames through multi-level surface alignment, UMA~\cite{zhu2025ultra} significantly improves visual fidelity and achieves the state-of-the-art performance.
Despite their high visual fidelity, these methods incur substantial computational or memory overhead due to excessive 2D convolutions on high-resolution, multi-channel feature maps.

\subsection{Efficient Animatable Avatars}
To address the high computational and memory demands of photorealistic dynamic avatars, recent works focus on improving their efficiency under limited resource budgets.
Pixel Codec Avatars~\cite{Ma_2021_CVPR} and MoRF~\cite{bashirov2024morf} adopt neural deferred rendering, but suffer from low rendering resolution and limited frame rates on mobile devices. 
SplattingAvatar~\cite{SplattingAvatar:CVPR2024} achieves real-time performance on mobile devices, though it does not model pose-dependent effects.
SqueezeMe~\cite{iandola2025squeezeme} presents a highly efficient formulation by distilling pose-dependent Gaussian correctives into a linear representation, enabling real-time inference on mobile devices.
This strong linearization, however, limits motion-dependent dynamics, leading to less diverse wrinkle patterns and their method struggles to model large-scale non-rigid deformations of loose garments such as skirts.
TaoAvatar~\cite{chen2025taoavatar} adopts a two-stage representation, modeling coarse canonical offsets with an MLP and Gaussian correctives with learned blend-shape bases distilled from Animatable Gaussians~\cite{li2024animatable}.
Specifically, the Gaussian correctives are densely defined over all Gaussian splats and parameterized by a compact set of learned bases with pose-dependent coefficients.
While efficient, this global formulation of the Gaussian correctives introduces two key limitations:
First, it limits the expressiveness of the model, making it difficult to capture complex deformations under diverse motions.
Second, although increasing the number of bases can partially improve expressiveness, it substantially increases the model size and still struggles to capture high-frequency dynamics, often leading to blurred appearance under non-standing poses.

In contrast, under real-time and mobile constraints, our method enables high-fidelity animation and rendering of full-body Gaussian avatars that capture rich motion-dependent geometric and appearance details, achieving comparable or even superior fidelity to server-based models (Sec.~\ref{sec:results}).
\section{Preliminaries} 
\label{sec:preliminaries}
%
%%%%
In this section, we first introduce our data assumptions (Sec.~\ref{subsec:data_asumption}), followed by an overview of the teacher model (Sec.~\ref{subsec:teacher}).
Finally, we analyze the limitations of the teacher model (Sec.~\ref{subsec:limitation}), which motivates our method.
%
%
%%%%%%%%%%%%%%%%%%%%%%%%%%%%%%%%%%%%%%
%
\begin{figure*}[tp]
    \centering
    \includegraphics[width=\linewidth]{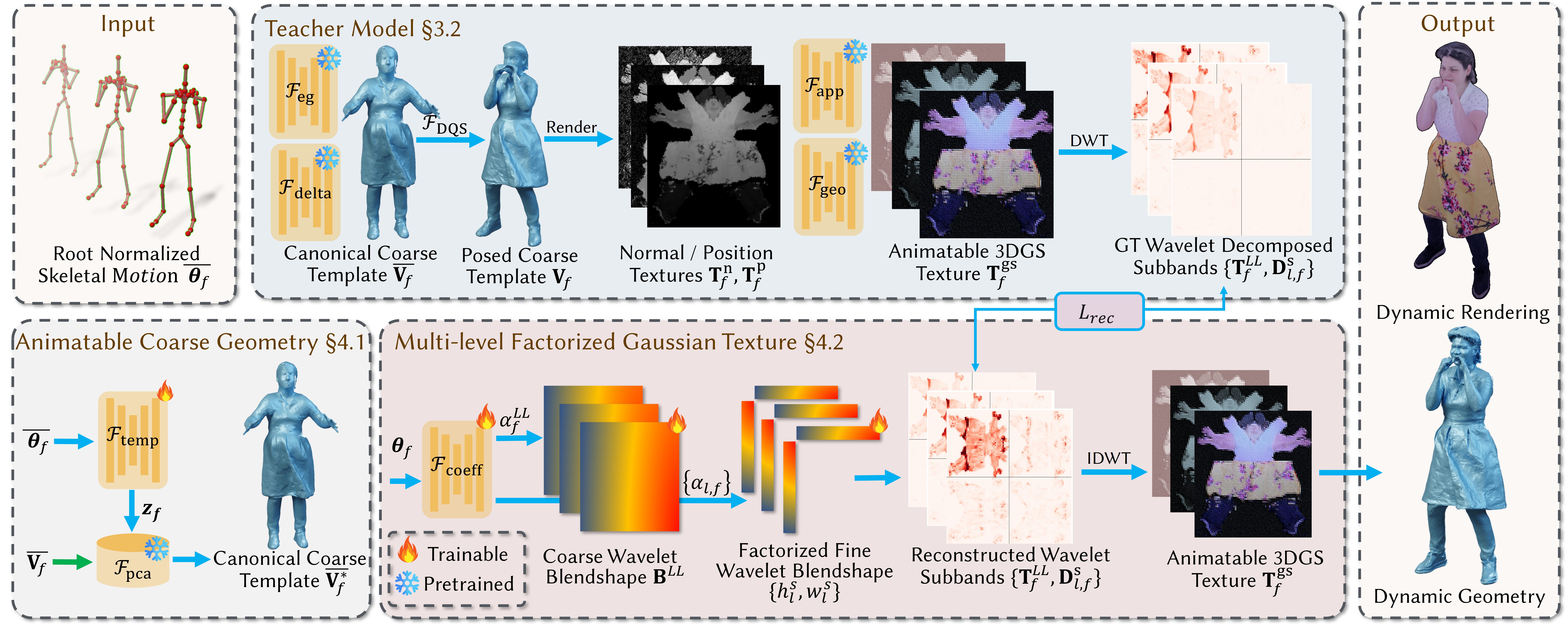}
    \vspace{-15pt}
    \caption{
       \textbf{Overview.} Given root-normalized skeletal motion $\bar{\boldsymbol{\theta}}_f$ as input, we first train a teacher model that models the coarse geometry with a template mesh $\bar{\mathbf{V}}_{f}$ and fine geometry and appearance with 3D Gaussian splat textures $\mathbf{T}^{\mathrm{gs}}_f$.
        We further decompose $\mathbf{T}^{\mathrm{gs}}_f$ with a wavelet transform to obtain multi-level supervision for distillation.
        To derive a compact, mobile-ready representation, we model the \textbf{coarse geometry} $\bar{\mathbf{V}}_{f}$ in a PCA subspace defined in canonical space, with the coefficients predicted by a lightweight MLP $\mathcal{F}_{\mathrm{temp}}$.
        For high-frequency geometry and appearance, we propose \textbf{Wavelet-guided Multi-level Factorized Gaussian Textures}, which represent the animatable avatar with structured blendshapes under a significantly reduced computational budget.
    }
    \label{fig:pipeline}

\end{figure*}
%
%%%%%%%%%%%%%%%%%%%%%%%%%%%%%%%%%%%%%%
%
%
\subsection{Data Assumption} 
\label{subsec:data_asumption}
We assume a rigged and skinned template mesh $\bar{\mathbf{V}} \in \mathbb{R}^{V \times 3}$, segmented multi-view videos $\mathbf{I}_{f,c} \in \mathbb{R}^{H \times W}$, where $f$ and $c$ denote the frame and camera indices, respectively. $W$ and $H$ denote the width and height of multi-view captured images.
Each frame $\mathbf{I}_{f,c}$ is annotated with the camera calibrations $\mathbf{C}_c$ and a 3D skeletal pose $\boldsymbol{\theta}_f \in \mathbb{R}^D$ estimated via a commercial marker-less motion capture system~\cite{captury}, where $D$ denotes the skeletal degrees of freedom (DoFs).
Moreover, we provide per-frame ground-truth 3D geometry $\mathbf{V}^\mathrm{gt}_f$ reconstructed from multi-view videos using NeuS2~\cite{neus2}, which balances reconstruction speed and quality.
To capture temporal motion context, we define a motion descriptor $\boldsymbol{\bar{\theta}}_{f} \in \mathbb{R}^{k \times D}$ as the skeletal poses within a sliding window spanning frames $f - k + 1$ to frame $f$ with root joint translations normalized relative to frame $f$.

\subsection{Teacher Model: Animatable Gaussian Texture} 
\label{subsec:teacher}
We adopt UMA~\cite{zhu2025ultra} as the teacher model for its strong capability in modeling both large-scale clothing dynamics and fine-grained wrinkle details.
Specifically, UMA represents motion-dependent deformations using a \textbf{drivable human template mesh} with motion-conditioned vertex offsets, and models high-frequency geometry and appearance via \textbf{motion-aware Gaussian splats} in the UV space of the template mesh.

\par \noindent \textbf{Drivable Human Template Mesh.} 
We employ a drivable human template mesh $\mathbf{V}_{f}$ to model motion-aware, large-scale clothing deformations (e.g., skirt swaying) conditioned on the motion descriptor $\boldsymbol{\bar{\theta}}_{f}$
%%%%%
\begin{align}\mathbf{V}_{f} \label{eq:template}
&=f_\mathrm{dq}(\bar{\mathbf{V}}_{f},\boldsymbol{\theta}_f) \\
&= f_\mathrm{dq}(f_\mathrm{eg}(\mathbf{A}_f, \mathbf{T}_f,\bar{\mathbf{V}})+ \boldsymbol{\delta}_f,\boldsymbol{\theta}_f) \\
&= f_\mathrm{dq}(f_\mathrm{eg}(\mathcal{F}_\mathrm{eg}(\bar{\boldsymbol{\theta}}_{f}),\bar{\mathbf{V}})+ \mathcal{F}_\mathrm{delta}(\bar{\boldsymbol{\theta}}_{f}),\boldsymbol{\theta}_f),
\end{align}
%%%%%
where $f_\mathrm{dq}(\cdot,\boldsymbol{\theta}_f)$ denotes Dual Quaternion skinning~\cite{kavan2007skinning}, and $\bar{\mathbf{V}}_{f}$ represents the canonical template mesh with motion-aware deformations.
$\mathbf{A}_f \in \mathbb{R}^{V \times 4}$ and $\mathbf{T}_f \in \mathbb{R}^{V \times 4}$ denote the rotation and translation quaternions of the embedded graph nodes, and $\boldsymbol{\delta}_f \in \mathbb{R}^{V \times 3}$ represents the per-vertex offsets that capture finer wrinkles in the canonical space.
Specifically, following the design by Habermann et al.~\cite{habermann2021}, the rotation $\mathbf{A}_f$ and translation quaternions $\mathbf{T}_f$, together with per-vertex displacement $\boldsymbol{\delta}_f$, are predicted by separate graph convolutional neural networks, namely, embedded deformation network $f_\mathrm{eg}$ and offset network $f_\mathrm{delta}$, which are trained by minimizing
\begin{equation}
\mathcal{L}_\mathrm{cham}(\mathbf{V}_f, \mathbf{V}^\mathrm{gt}_f) + \mathcal{L}_\mathrm{spatial}(\mathbf{V}_f) +  \mathcal{L}_\mathrm{cor-vrt}(\mathbf{V}_f, \mathbf{V}^\mathrm{gt}_f)
\end{equation}
where $\mathcal{L}_\mathrm{cham}$ and $\mathcal{L}_\mathrm{spatial}$ denote the Chamfer distance and a spatial regularization term.
$\mathcal{L}_\mathrm{cor-vrt}$ is a vertex alignment term introduced in UMA~\cite{zhu2025ultra}, which enforces correspondence between the predicted and ground-truth surfaces through a foundational point tracker~\cite{karaev2024cotracker}.

\par \noindent \textbf{Motion-aware Animatable Gaussian Splats.} 
On top of the drivable template mesh, UMA~\cite{zhu2025ultra} models fine-scale geometry and dynamic appearance using motion-aware Gaussian splats $\mathbf{T}^{\mathrm{gs}}_f =(\boldsymbol{\bar{\mu}}^{\mathrm{uv}}_i, \mathbf{\bar{d}}^{\mathrm{uv}}_i, \mathbf{q}^{\mathrm{uv}}_i, \mathbf{s}^{\mathrm{uv}}_i, \mathbf{\alpha}^{\mathrm{uv}}_i, \boldsymbol{\eta}^{\mathrm{uv}}_i)_f \in \mathbb{R}^{N\times26}$ in the texture space of the template mesh.
Here, $\boldsymbol{\bar{\mu}}^{\mathrm{uv}}_i$ denotes the canonical Gaussian positions obtained via barycentric interpolation on the non-rigidly deformed template $\bar{\mathbf{V}}_f$, $\mathbf{\bar{d}}^{\mathrm{uv}}_i$ denotes the motion-aware offsets for capturing fine wrinkles, and $\boldsymbol{\eta}^{\mathrm{uv}}_i$ denotes the appearance coefficients represented with degree-1 spherical harmonics.
Specifically, UMA adopts two UNets~\cite{ronneberger2015u}, $\mathcal{E}_{\mathrm{geo}}$ and $\mathcal{E}_{\mathrm{app}}$, to predict the geometric attributes $(\mathbf{\bar{d}}^{\mathrm{uv}}_i, \mathbf{q}^{\mathrm{uv}}_i, \mathbf{s}^{\mathrm{uv}}_i, \mathbf{\alpha}^{\mathrm{uv}}_i)_f$ and appearance attributes $(\boldsymbol{\eta}^{\mathrm{uv}}_i)_f$ of the 3D Gaussian splats from the positional and normal textures $(\mathbf{T}^{\mathrm{p}}_f, \mathbf{T}^{\mathrm{n}}_f)$ rendered from the posed template mesh $\mathbf{V}_f$.
Both networks are supervised using multi-view images with
\begin{equation} \label{eq:loss_gauss}
\small
\mathcal{L}_{1}(I^\mathrm{R}_{f,c}, \mathbf{I}_{f,c}) + \mathcal{L}_\mathrm{ssim}(I^\mathrm{R}_{f,c}, \mathbf{I}_{f,c}) + \mathcal{L}_\mathrm{mrf}(I^\mathrm{R}_{f,c}, \mathbf{I}_{f,c}) + \mathcal{L}_\mathrm{cor-tex},
\end{equation}
where $I^\mathrm{R}_{f,c}$ denotes the rendered image, $\mathcal{L}_\mathrm{mrf}$ indicates the perceptual loss~\cite{wang2018image}, and $\mathcal{L}_\mathrm{cor-tex}$ is a tracking-based texel alignment term introduced in UMA~\cite{zhu2025ultra}.
\subsection{Limitations and Motivations} 
\label{subsec:limitation}
Despite its strong ability to capture dynamic wrinkle details and appearance, UMA~\cite{zhu2025ultra}, like other high-quality 3D Gaussian-based avatar approaches, remains both memory- and computation-intensive.
This hinders its deployment on consumer-grade hardware, including PCs with consumer-grade GPUs and VR headsets, especially under real-time constraints.
We argue that these limitations mainly arise from dense \emph{2D feature processing} with \emph{large spatial resolution}, \emph{many feature channels}, and \emph{multiple convolutional blocks}.

\noindent \textbf{Memory Bottleneck.} 
Methods such as UMA~\cite{zhu2025ultra} and Animatable Gaussians~\cite{li2024animatable} perform dense 2D convolutions over high-resolution feature maps (e.g., $768 \times 768$ and $512 \times 512$) with a large number of channels, leading to high-dimensional intermediate activations throughout the network.
As these activations must be repeatedly stored and accessed across multiple layers, they incur substantial memory footprints and heavy memory bandwidth consumption, which poses a major challenge for deployment on resource-constrained devices, especially standalone VR platforms.

\noindent \textbf{Computational Bottleneck.} 
Since the cost of standard 2D convolution scales approximately with $H \times W \times C_{in} \times C_{out} \times K^2$ per layer, repeatedly processing high-resolution, multi-channel feature maps across multiple convolutional blocks becomes prohibitively expensive for real-time deployment on consumer-grade and mobile devices. 

Based on these observations, we identify two key design goals for mobile-friendly animatable avatars while preserving the quality of the teacher model: (1) minimizing computation on high-resolution, multi-channel 2D feature maps, and (2) reducing reliance on deep 2D convolutional pipelines.

\section{Methodology} 
\label{sec:method}
In this section, we introduce our approach for distilling an ultra-high-quality animatable clothed human avatar from the teacher model into a compact and efficient representation. 
First, we present the distillation of an animatable clothed human template mesh that serves as the coarse geometric proxy (Sec.~\ref{subsec:coarsegeodistall}).
Next, we introduce an efficient and compactly factorized representation for modeling the motion-dependent detailed Gaussian splat geometry and appearance of the animatable human avatar (Sec.~\ref{subsec:finegeodistill}).
Finally, we describe how this compact representation is learned to distill both dynamic geometry and motion-aware appearance from the teacher model (Sec.~\ref{subsec:supervsion}).

\subsection{Animatable Coarse Geometry}
\label{subsec:coarsegeodistall}
UMA~\cite{zhu2025ultra} demonstrates that a well-tracked drivable template mesh with consistent vertex correspondence provides a strong foundation for learning detailed geometry and appearance. 
However, the multi-block graph convolution networks~\cite{habermann2021} for modeling the detailed templates $F_\mathrm{eg}$ and $F_\mathrm{delta}$ remain computationally expensive and are difficult to deploy in resource-constrained settings.

To obtain a lightweight representation while preserving the vertex correspondence of the coarse template mesh $\bar{\mathbf{V}}_{f}$ of UMA~\cite{zhu2025ultra}, we construct a PCA subspace $\mathcal{F}_{\mathrm{pca}}(\cdot)$~\cite{mackiewicz1993principal} over the canonically deformed mesh vertex positions $\bar{\mathbf{V}}_{f}$. 
For a given motion descriptor $\bar{\boldsymbol{\theta}}_f$, a shallow MLP $\mathcal{F}_{\mathrm{temp}}(\cdot)$ first predicts the PCA coefficients $\mathbf{z}_{f}$, which are then decoded through the PCA subspace to reconstruct the coarse template mesh $\bar{\mathbf{V}}^{\star}$ in the canonical space:
\begin{align}
\bar{\mathbf{V}}^{\star} = \mathcal{F}_{\mathrm{pca}}(\mathbf{z}_{f}) = \mathcal{F}_{\mathrm{pca}}(\mathcal{F}_{\mathrm{temp}}(\bar{\boldsymbol{\theta}}_f)),
\end{align}
where we empirically set the number of PCA coefficients to $128$, which is sufficient to capture the coarse template deformations across roughly 20k training frames with diverse motions.
We then apply dual-quaternion skinning with the target pose $\boldsymbol{\theta}_f$ to obtain the final posed mesh:
\begin{align}
\mathbf{V}^{\star} &= f_\mathrm{dq}(\bar{\mathbf{V}}^{\star}, \boldsymbol{\theta}_f).
\end{align}

While this PCA formulation is effective for coarse template meshes with only a few thousand vertices, it does not scale well to modeling fine-grained human surface details, as capturing dynamic detailed geometry and appearance typically requires hundreds of thousands of Gaussian splats (e.g., $\sim$300K).
As shown in Tab.~\ref{tab:ablations}, applying PCA to such high-dimensional signals would require an excessively large basis, making it impractical for deployment on personal computers and VR headsets.
In other words, directly replacing dense 2D convolutions with a single global PCA basis over the full set of Gaussian splats does not fundamentally resolve the memory and computation bottlenecks.

%
%%%%%%%%%%%%%%%%%%%%%%%%%%%%%%%%%%%%%%
%
\begin{figure}[t]
    \centering
    \includegraphics[width=\linewidth]{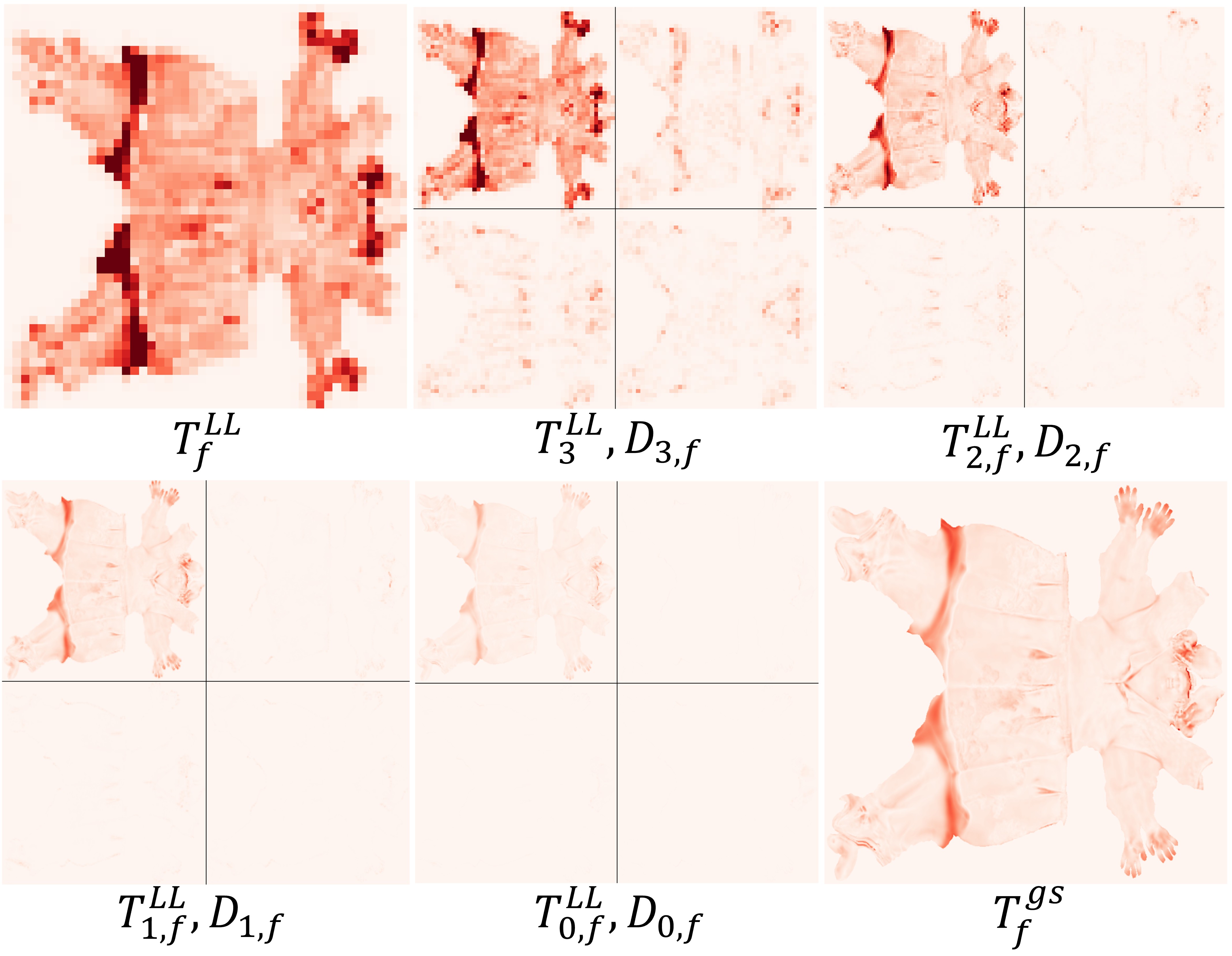}
    \vspace{-15pt}
    \caption{
        \textbf{Intuition.} The proposed \textbf{Wavelet-guided Multi-level Factorized Gaussian Texture} is based on the observation that different wavelet subbands of the Gaussian texture $\mathbf{T}^{\mathrm{gs}}_f$ exhibit distinct structural properties.
        The coarsest low-frequency subband $\mathbf{T}^{\mathrm{LL}}_{f}$ contains most of the signal energy but has a low spatial resolution.
        The intermediate detail subbands $\mathbf{D}_{l,f}$, $l \in \{2,3\}$, are sparse and thus well suited for 1D factorized modeling.
        The finest detail subbands $\mathbf{D}_{l,f}$, $l \in \{0,1\}$, contain only limited energy and can therefore be approximated by the average over training frames $\bar{\mathbf{D}}_l$.
    }
    \label{fig:intuition}
\end{figure}
%
%%%%%%%%%%%%%%%%%%%%%%%%%%%%%%%%%%%%%%
%

\subsection{Wavelet-guided Multi-level Factorized Gaussian Texture}
\label{subsec:finegeodistill}
As discussed in Sec.~\ref{subsec:limitation}, dense 2D convolutions over high-resolution feature maps are the major bottleneck in terms of memory and computation.
%%%%%%%%%%%%%%%%%%%%%%%%%%%%%%%%%%%%%%%%%%%%%%%%%%%%%%%%%%%%%%%%%%%%%%%%%%%%%%%%%%%%%%%%%%%%%%%%%%%%%%%
% TaoAvatar is already in the natrual alternaltive, 
%%%%%%%%%%%%%%%%%%%%%%%%%%%%%%%%%%%%%%%%%%%%%%%%%%%%%%%%%%%%%%%%%%%%%%%%%%%%%%%%%%%%%%%%%%%%%%%%%%%%%%%
A vanilla alternative is to replace the deep 2D convolutional blocks with a learned blendshape formulation over the full set of motion-dependent Gaussian splats, as explored in TaoAvatar~\cite{chen2025taoavatar}.
However, directly learning blendshape-style correctives over the full set of Gaussian splats fails to faithfully model complex motion-dependent Gaussian splat attributes, as memory and computation overhead already become substantial before sufficient expressive power is reached (Tab.~\ref{tab:quantitive}). 

\noindent \textbf{Single-level Spatial Factorization.}
Recent advances in low-rank decomposition~\cite{hu2022lora} and factorized neural representations~\cite{chen2022tensorf}, suggest that high-dimensional spatial signals or tensors can be compactly modeled through factorization.
Inspired by this idea, we introduce a first baseline that represents the full-resolution 2D animatable Gaussian texture attributes $\mathbf{T}^{\mathrm{gs}}_f(:,:,c)$, as a weighted sum of outer products between learned 1D bases:
\begin{align}
\mathbf{T}^{\mathrm{gs}}_f%(:,:,c)
&=
\bar{\mathbf{T}}^{\mathrm{gs}}_f + 
\sum_{r=1}^{R}
\alpha_f(r,c)\,
\mathbf{h}_f(:,r,c)\otimes \mathbf{w}_f(:,r,c),
\end{align}
where $\mathbf{h}_f \in \mathbb{R}^{H \times R \times C}$ and $\mathbf{w}_f \in \mathbb{R}^{W \times R \times C}$ denote the learned 1D bases along the height and width axes, respectively.
$\alpha_f \in \mathbb{R}^{R \times C}$ denotes the motion-dependent blendshape coefficients, and $R$ denotes the number of blendshape bases.
Following the coarse geometry modeling in Sec.~\ref{subsec:coarsegeodistall}, we can similarly predict the blendshape coefficients $\alpha_f$ from the skeletal motion descriptor $\bar{\boldsymbol{\theta}}_f$ as:
\begin{align}
\alpha_f = \mathcal{F}_{\mathrm{bs}}(\bar{\boldsymbol{\theta}}_f),
\end{align}
where $\mathcal{F}_{\mathrm{bs}}(\cdot)$ denotes a shallow MLP.

Compared with directly learning blendshapes on full-resolution Gaussian textures, as in TaoAvatar~\cite{chen2025taoavatar}, the factorized formulation substantially reduces both memory and computational costs.
However, we find that directly applying low-rank decomposition in the spatial domain of the full-resolution Gaussian textures remains suboptimal, since low- and high-frequency components are entangled, making the signal less compatible with the low-rank and sparsity assumptions required for effective factorization.
As a result, a larger rank $R$ is required for faithful reconstruction, which undermines the efficiency and memory gains of the factorized formulation, also making it hard to learn the detailed geometry and appearance (Sec.~\ref{subsec:ablations}).

\noindent \textbf{Wavelet-guided Multi-level Factorized Blendshapes.}
Effective low-rank decomposition relies on the target signal exhibiting sufficient low-rank structure and sparsity.
Motivated by this, we transform the Gaussian texture map $\mathbf{T}^{\mathrm{gs}}_f$ into the wavelet domain, where motion-dependent low-frequency components are naturally separated from high-frequency details, yielding a representation that better satisfies these structural assumptions.
Specifically, we apply a multi-level 2D discrete wavelet transform (DWT) to $\mathbf{T}^{\mathrm{gs}}_f$ using a biorthogonal wavelet basis.
At each level $l$, the wavelet transform recursively decomposes the signal into one low-frequency component $\mathbf{T}^{\mathrm{LL}}_{l,f}$ and three high-frequency detail subbands $\mathbf{D}_{l,f}$:
\begin{equation}
\mathrm{DWT}^{(l)}(\mathbf{T}^{\mathrm{gs}}_{l,f})
=
\left\{
\mathbf{T}^{\mathrm{LL}}_{l,f},\;
\mathbf{D}_{l,f}=\{\mathbf{T}^{\mathrm{LH}}_{l,f},\mathbf{T}^{\mathrm{HL}}_{l,f},\mathbf{T}^{\mathrm{HH}}_{l,f}\}
\right\},
\end{equation}
where we use a four-level decomposition with levels $l \in \{0,1,2,3\}$, yielding the coarsest low-frequency subband $\mathbf{T}^{\mathrm{LL}}_{f} = \mathbf{T}^{\mathrm{LL}}_{3,f}  \in \mathbb{R}^{48 \times 48 \times C}$.

As shown in Fig.~\ref{fig:intuition}, the resulting subbands exhibit distinct structural characteristics across levels, which motivates using different representations for different frequency bands:

\paragraph{Dynamic low-frequency subband}
The coarsest low-frequency subband $\mathbf{T}^{\mathrm{LL}}$ preserves most of the motion-aware information in the dynamic Gaussian texture $\mathbf{T}^{\mathrm{gs}}_f$, while significantly reducing the spatial resolution.
Although $\mathbf{T}^{\mathrm{LL}}_{l,f}$ remains relatively dense and is, thus, less suitable for spatial factorization, its low spatial dimensionality makes it tractable to directly model pose-dependent variations using blendshape-based formulation without incurring substantial memory or computational overhead:
\begin{align}
\mathbf{T}^{\mathrm{LL}}_{f}%(:,:,c)
&=
\bar{\mathbf{T}}^{\mathrm{LL}}%(:,:,c)
+
\sum_{r=1}^{R}
\alpha^{\mathrm{LL}}_f(r,c)\,
\mathbf{B}^{\mathrm{LL}}(:,:,r,c),
\end{align}
where $\bar{\mathbf{T}}^{\mathrm{LL}} \in \mathbb{R}^{H_3 \times W_3 \times C}$ denotes the canonical low-frequency subband, and
$\mathbf{B}^{\mathrm{LL}} \in \mathbb{R}^{H_3 \times W_3 \times R \times C}$ denotes the learned 2D blendshape bases.
$\alpha^{\mathrm{LL}}_f \in \mathbb{R}^{R \times C}$ denotes the motion-dependent blendshape coefficients, where $R$ is the number of blendshape bases.
The coefficients are predicted from the skeletal motion descriptor $\bar{\boldsymbol{\theta}}_f$:
\begin{align}
\alpha^{\mathrm{LL}}_f = \mathcal{F}_{\mathrm{coeff}}(\bar{\boldsymbol{\theta}}_f).
\end{align}

While the coarsest subband $\mathbf{T}^{\mathrm{LL}}_{f}$ could also be predicted using 2D convolutional blocks, as in prior high-fidelity avatar approaches~\cite{li2024animatable,Pang_2024_CVPR,zhu2025ultra}, we find direct blendshape modeling more effective when the computation and memory budget is restricted (Sec.~\ref{subsec:ablations}).

\paragraph{Dynamic mid-frequency subbands}
For the intermediate wavelet decomposition levels ($l \in \{2,3\}$), the detail subbands $\mathbf{D}_{l,f}$ exhibit notable sparsity while still retaining meaningful pose-dependent geometric variations. 
This sparsity makes them well suited to separable outer-product factorization, 
where each detailed subband $\mathbf{T}^{s}_{l,f}$ is modeled as a weighted sum of outer products of 1D basis components along the height and width axes:
\begin{align}
\mathbf{T}^{s}_{l,f}%(:,:,c)
&=
\bar{\mathbf{T}}^{s}_{l}%(:,:,c)
+
\sum_{r=1}^{R_l}
\alpha_{l,f}(r,c)\,
\mathbf{h}^{s}_{l}(:,r,c)\otimes \mathbf{w}^{s}_{l}(:,r,c),
\end{align}
where $s \in \{\mathrm{LH},\mathrm{HL},\mathrm{HH}\}$ and $l \in \{2,3\}$.
$\bar{\mathbf{T}}^{s}_{l} \in \mathbb{R}^{H_l \times W_l \times C}$ denotes the average subband of orientation $s$ at level $l$.
$\mathbf{h}^{s}_{l} \in \mathbb{R}^{H_l \times R_l \times C}$ and $\mathbf{w}^{s}_{l} \in \mathbb{R}^{W_l \times R_l \times C}$ denote the subband-specific 1D basis components along the height and width axes, respectively.
$\alpha_{l,f} \in \mathbb{R}^{R_l \times C}$ denotes the motion-dependent coefficients shared by the three detail subbands at level $l$, where $R_l$ is the number of basis components used at that level.
In practice, the coefficients of all intermediate detail subbands are jointly predicted from the skeletal motion descriptor $\bar{\boldsymbol{\theta}}_f$ by a single shared MLP, together with the coefficients from the coarsest level:
\begin{align}
\alpha^{\mathrm{LL}}_f, \left\{\alpha_{l,f}\right\}_{l \in \{2,3\}}
=
\mathcal{F}_{\mathrm{coeff}}(\bar{\boldsymbol{\theta}}_f).
\end{align}

\paragraph{High-frequency subbands}
For the finer decomposition levels ($l \in \{0,1\}$), the remaining signal energy is negligible. %\sun{illustrate with figure}
% all in Fig.3
We therefore approximate the remaining detail subbands $\mathbf{D}_{l,f}$ at these levels using their channel-wise spatial mean computed over all training frames:
\begin{equation}
\mathbf{D}_{l,f} = 
\bar{\mathbf{D}}_l
=
\frac{1}{N}\sum_{f=1}^{N}\mathbf{D}_{l,f}.
\end{equation}

During inference, the averaged subbands $\bar{\mathbf{D}}_0$ and $\bar{\mathbf{D}}_1$ 
are precomputed offline and treated as constants.
Owing to the linearity of the IDWT, their contribution to the final reconstruction 
can be precomputed as a static offset $\bar{\mathbf{T}}^{\mathrm{static}}$, 
reducing per-frame reconstruction to a partial IDWT over the dynamic subbands 
$\{\mathbf{T}^{\mathrm{LL}}_3, \mathbf{D}_{3,f}, \mathbf{D}_{2,f}\}$ followed 
by a single addition:
\begin{equation}
\mathbf{T}^{\mathrm{gs}}_f
=
\underbrace{\mathrm{IDWT}
\big(
\mathbf{T}^{\mathrm{LL}}_3,\,
\mathbf{D}_{3,f},\,
\mathbf{D}_{2,f},\,
\mathbf{0},\,
\mathbf{0}
\big)}_{\text{per-frame}}
+
\underbrace{\bar{\mathbf{T}}^{\mathrm{static}}}_{\text{precomputed}},
\end{equation}
where $\bar{\mathbf{T}}^{\mathrm{static}} = \mathrm{IDWT}(\mathbf{0}, \mathbf{0}, \mathbf{0}, \bar{\mathbf{D}}_1, \bar{\mathbf{D}}_0)$.
Since the detailed subbands at levels 0 and 1 are zeroed out, the IDWT at the two highest spatial resolutions ($384{\times}384$ and $768{\times}768$) incurs only one quarter of the original cost, because the standard four-branch reconstruction reduces to a single low-pass branch.

\par
Taken together, these design choices yield several efficiency advantages: The blendshape-like formulation avoids dense 2D convolutions over high-resolution feature maps, substantially reducing per-frame computation and memory bandwidth. 
Meanwhile, the wavelet-guided multi-level decomposition further improves parameter efficiency by assigning compact, structure-aware representations to each frequency band, enabling a favorable quality–efficiency trade-off particularly suited for resource-constrained devices.

\subsection{Supervision and Training Objectives}
\label{subsec:supervsion}
Based on the proposed representation, namely the animatable coarse geometry (Sec.~\ref{subsec:coarsegeodistall}) and the multi-level factorized Gaussian texture (Sec.~\ref{subsec:finegeodistill}), we now introduce the supervision and training objectives used to distill the compact animatable avatar from the teacher model.

\noindent\textbf{Animatable Coarse Geometry.}
We supervise the coarse-geometry coefficient network $\mathcal{F}_{\mathrm{temp}}(\cdot)$ by minimizing the following loss
\begin{equation}
    \mathcal{L}_{\mathrm{coarse}} = \| \mathcal{F}_{\mathrm{temp}}(\bar{\boldsymbol{\theta}}_f) - \mathbf{z}^{\mathrm{gt}}_{f} \|_1,
\end{equation}
where $\mathbf{z}^{\mathrm{gt}}_{f}$ denotes the ground-truth PCA coefficients obtained by projecting the coarse template mesh $\bar{\mathbf{V}}_{f}$ generated by the teacher model onto the coarse-template PCA space $\mathcal{F}_{\mathrm{pca}}(\cdot)$.

\noindent\textbf{Factorized Gaussian Geometry and Appearance.}
We supervise both the geometry and appearance coefficient networks using the same subband reconstruction loss, which is applied to the wavelet subbands of their corresponding Gaussian attributes generated by the teacher model.
Specifically, we supervise the wavelet blendshapes and motion-aware blendshape coefficient networks, i.e., $(\mathbf{B}^{\mathrm{LL}}, \mathbf{h}^{s}_{l}, \mathbf{w}^{s}_{l})$ together with $\mathcal{F}_{\mathrm{coeff}}(\cdot)$, by minimizing
\begin{equation}
\small
\mathcal{L}_{\mathrm{rec}}
=
\lambda_{\mathrm{LL}}
\left\|
\mathbf{T}^{\mathrm{LL}}
-
\mathbf{T}^{\mathrm{LL},\mathrm{gt}}_{f}
\right\|_{1}
+
\sum_{l\in\{2,3\}}
\lambda_{l}
\left\|
\mathbf{D}_l
-
\mathbf{D}^{\mathrm{gt}}_{l,f}
\right\|_{1},
\end{equation}
where $\mathbf{T}^{\mathrm{LL},\mathrm{gt}}_{f}$ and $\mathbf{D}^{\mathrm{gt}}_{l,f}$ denote the ground-truth wavelet subbands obtained by decomposing the full-resolution Gaussian splat texture $\mathbf{T}^{\mathrm{gs}}_f$ generated by the teacher model.
$\lambda_{\mathrm{LL}}$,$\lambda_2$,$\lambda_3$ are set to $5.0$, $1.0$ and $1.0$, respectively.

Notably, unlike the teacher model~\cite{zhu2025ultra}, which predicts the Gaussian splat appearance SH coefficients $\boldsymbol{\eta}^{\mathrm{uv}}_{i,f}$ directly in the posed space, we first transform them into the canonical space by factoring out the skeletal transformation $\mathbf{R}^{\mathrm{tex}}_{i,f}$, and then learn the canonical SH offset with respect to the canonical frame:
\begin{equation}
   \bar{\boldsymbol{\eta}}^{\mathrm{uv}}_{i,f}
   =
   \bar{\boldsymbol{\eta}}^{\mathrm{uv}}_{i}
   +
   \underbrace{
   \mathcal{W}((\mathbf{R}^{\mathrm{tex}}_{i,f})^{-1})\boldsymbol{\eta}^{\mathrm{uv}}_{i,f}
   -
   \bar{\boldsymbol{\eta}}^{\mathrm{uv}}_{i}
   }_{\text{canonical SH offset}},
\end{equation}
where $\bar{\boldsymbol{\eta}}^{\mathrm{uv}}_{i}$ denotes the SH coefficients of the canonical (approximately A-posed) frame, and $\mathcal{W}((\mathbf{R}^{\mathrm{tex}}_{i,f})^{-1})$ denotes the Wigner D-matrix that rotates the SH coefficients of each texel back into the canonical pose. 
When rendering the character in novel pose, instead of explicitly rotating the SH coefficients with the Wigner D-matrix, which is expensive to construct at runtime, we exploit the equivalent formulation of rotating the viewing direction from the posed space back into the canonical space and then evaluating the SH:
\begin{equation}
    c_{\mathrm{col},i}
    =
    SH\!\left((\mathbf{R}^{\mathrm{tex}}_{i,f})^{-1}\mathbf{d}, \bar{\boldsymbol{\eta}}^{\mathrm{uv}}_{i,f}\right),
\end{equation}
where $SH(\cdot,\cdot)$ denotes the function that converts the SH coefficients into RGB color $ c_{\mathrm{col},i}$, and $\mathbf{d}$ denotes the camera ray direction in world space.

%
%%%%%%%%%%%%%%%%%%%%%%%%%%%%%%%%%%%%%%%%%%%%
%
\section{Results} \label{sec:results}

%
%%%%%%%%%%%%%%%%%%%%%%%%%%%%%%%%%%%%%%
%
\begin{figure*}[tp]
    \centering
\includegraphics[width=\linewidth]{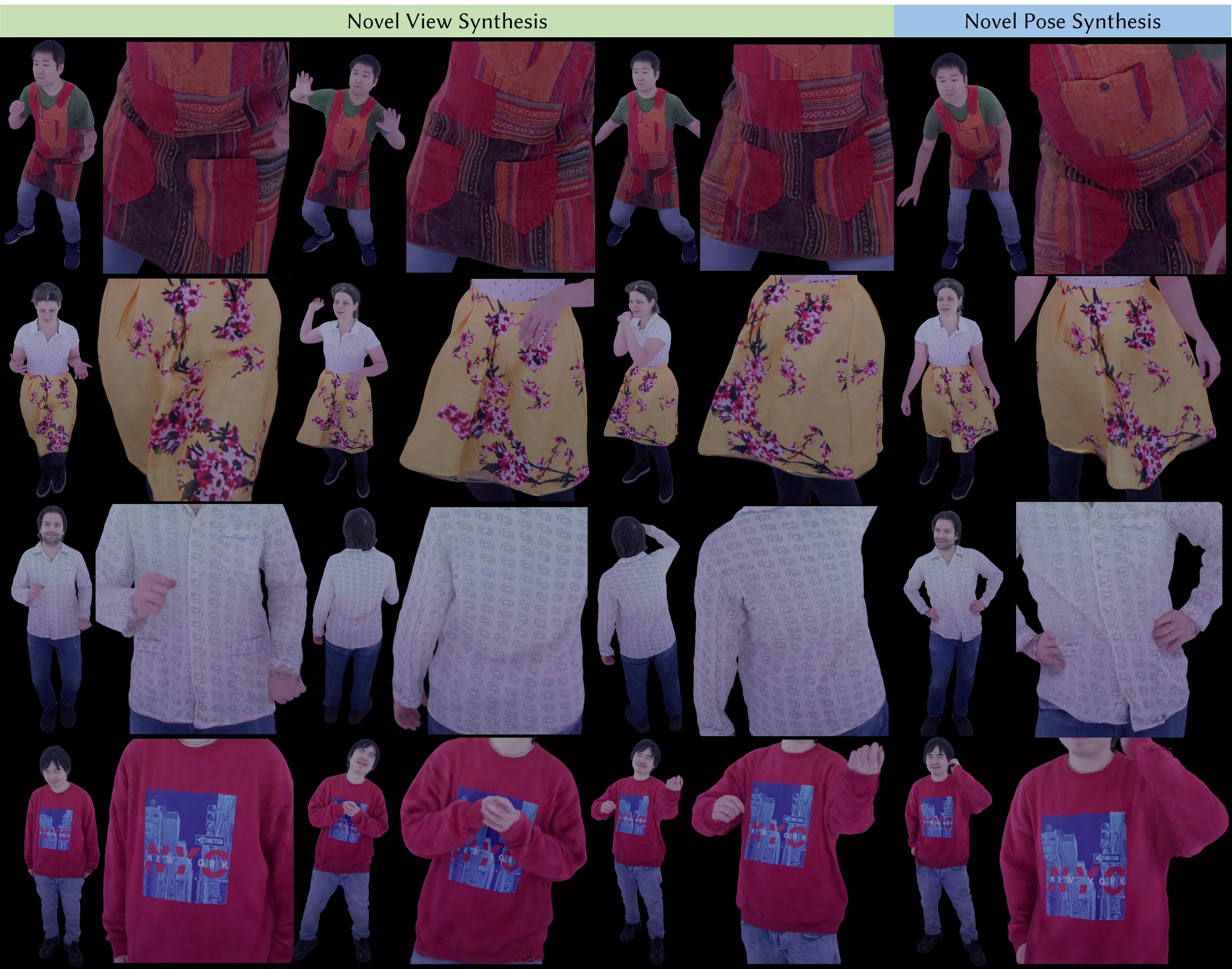}
    \vspace{-15pt}
    \caption{
        \textbf{Qualitative Rendering Results.} Given limited computation and memory budget, \textbf{MUA} produces detailed rendering with motion aware appearance and wrinkles.
        Please \textbf{zoom-in} to better observe the details.
    }
    \label{fig:gallray}
\end{figure*}
%
%%%%%%%%%%%%%%%%%%%%%%%%%%%%%%%%%%%%%%
%
%
%%%%%%%%%%%%%%%%%%%%%%%%%%%%%%%%%%%%%%
%
\begin{figure*}[tp]
    \centering
\includegraphics[width=\linewidth]{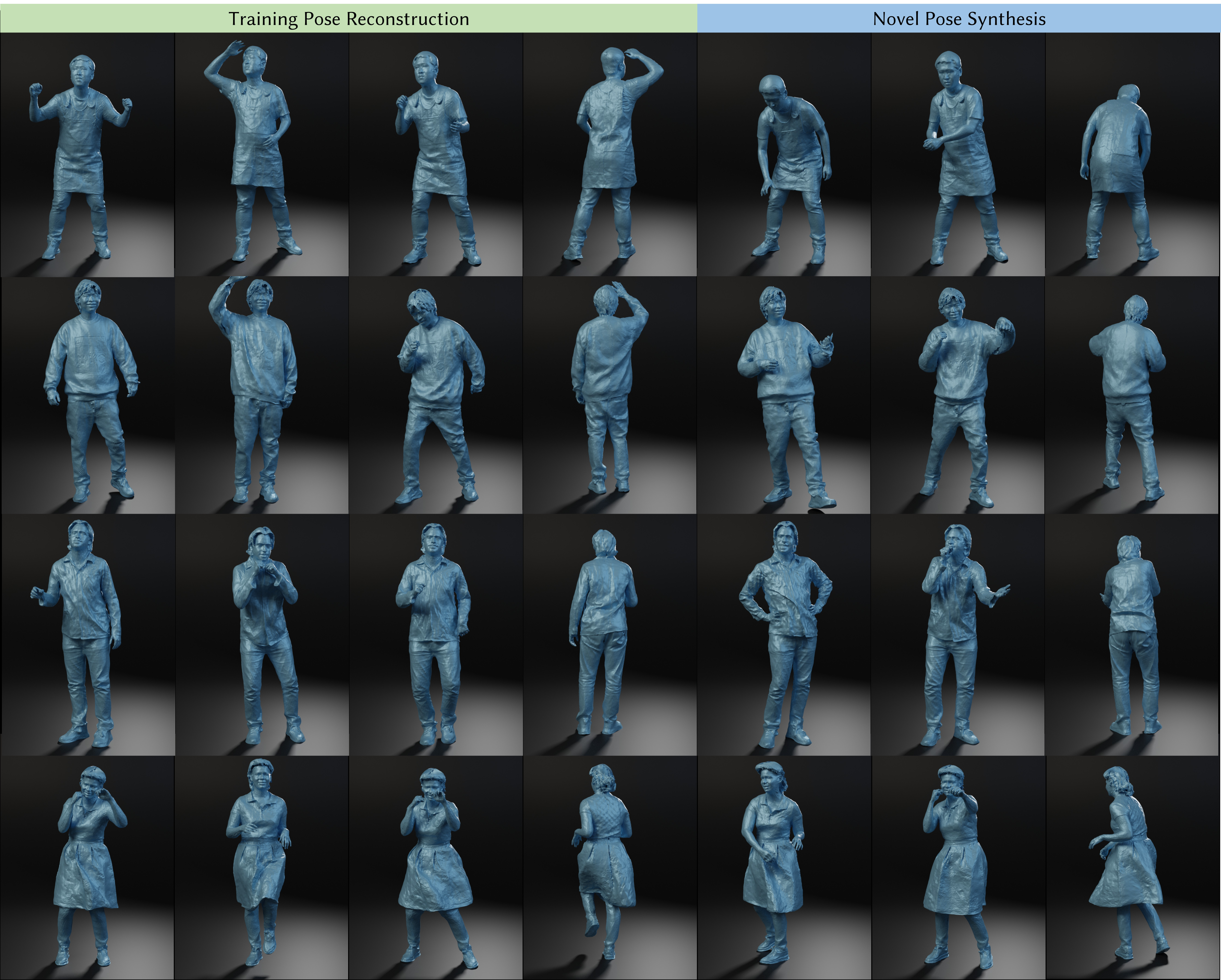}
    \vspace{-15pt}
    \caption{
        \textbf{Qualitative Geometry Results.} Given limited computation and memory budget, \textbf{MUA} and synthesize high-fidelity geometry with motion-aware wrinkles.
    }
    \label{fig:gallarygeo}
\end{figure*}
%
%%%%%%%%%%%%%%%%%%%%%%%%%%%%%%%%%%%%%%
%

\subsection{Implementation Details}
Our approach is implemented in PyTorch~\cite{paszke2017automatic}. 
For the coarse geometry stage, we train the motion-aware PCA coefficient network for 2 million iterations using the Adam optimizer~\cite{kingma2017adam} with a learning rate of $1\times10^{-4}$ and a batch size of 8, taking approximately $12$ hours. 
For the fine geometry and appearance stage, we set the numbers of blendshape bases for the coarsest subband $\mathbf{T}^{\mathrm{LL}}{f}$ and the detail subbands ${\mathbf{D}_{3,f}, \mathbf{D}_{2,f}}$ to $192$, $128$, and $96$ for geometry, and to $128$, $128$, and $96$ for appearance, respectively.
The learnable blendshapes and the coefficient MLPs are trained for 1 million iterations using the Adam optimizer~\cite{kingma2017adam} with a learning rate of $1\times10^{-4}$ and a batch size of 1, taking approximately $16$ hours.
For rendering, we adopt Analytical Splatting~\cite{liang2024analyticsplatting} for rasterizing the 3D Gaussian splats.
For all the stages, competing methods and ablation alternatives are trained and tested on a server with one NVIDIA H100 graphics card and an AMD EPYC 9554 CPU.
Moreover, to evaluate the efficiency and compactness of our approach, we report the FLOPs, model size, and runtime on a consumer-grade desktop PC equipped with an NVIDIA RTX 3090 GPU and an Intel i9-12900 CPU.

For the on-device demo, we implement the entire character inference pipeline in C$\#$ and GPU compute shaders in Unity3D, and adapt Gaussian splat rendering on top of the open-source Unity-VR-Gaussian-Splatting implementation~\cite{kleinbeck_multi-layer_2025}.
The resulting application is deployed on Meta Quest 3, where all the computation runs fully on-device at $24$ FPS.
We further support PC-connected mode via Meta Quest Link, where rendering is computed on a desktop PC and streamed to the Meta Quest 3 headset at $72$ or $90$ FPS.

\subsection{Dataset} 
We conduct all experiments on the dataset released by UMA~\cite{zhu2025ultra}.
The dataset comprises five subjects wearing garments with rich non-rigid clothing dynamics and intricate texture patterns.
For each subject, separate training and testing sequences are provided, covering a variety of everyday motions such as dancing, soccer kicking, and boxing.
The data is captured using a calibrated multi-view setup with $40$ synchronized cameras, each recording at a resolution of $3240 \times 6144$ and a frame rate of $25$ FPS.
The training and testing sequences contain roughly $17{,}000$ and $7{,}000$ frames, respectively.
Each frame is annotated with skeletal poses estimated by commercial 3D pose estimation software~\cite{captury}, foreground segmentations generated by Sapiens~\cite{khirodkar2024sapiens}, and pseudo ground-truth mesh geometry reconstructed with NeuS2~\cite{neus2}.
In addition, the dataset provides SMPL-X~\cite{SMPL-X:2019} parameters for all frames, which are used to train methods based on the SMPL-X parametric model, e.g., Animatable Gaussians~\cite{li2024animatable}.

\subsection{Metrics} 
We adopt the Peak Signal-to-Noise Ratio (PSNR) metric to measure the quality of the rendered image.
Besides, we adopt the Structural Similarity Index (SSIM) and learned perceptual image patch similarity (LPIPS)~\cite{zhang2018perceptual} that better mirrors human perception. 
Note that the metrics are computed at a resolution of $1620 \times 3072$, averaged over every 10th frame in the testing sequences, using two camera views that were excluded during training.
We measure computational complexity in Giga Floating-Point Operations (GFLOPs). 
To better reflect the complexity of the animatable avatar model itself, we exclude the computation associated with Gaussian splat rasterization when measuring the computational complexity.
Moreover, we report the parameter count of each model to reflect its memory footprint, i.e., Param (M), which can be a major constraint for deployment on mobile devices.
Lastly, we report the total runtime, including both avatar inference and Gaussian splat rendering, in frames per second (FPS) for each approach.

\subsection{Quantitative Results}
\textbf{Image Synthesis.} Fig.~\ref{fig:gallray} shows image synthesis results on the training split rendered in novel views, as well as on a standalone testing motion sequence.
Despite reducing computation by over 1000$\times$, and 10$\times$ fewer parameters compared with the teacher model, i.e., UMA~\cite{zhu2025ultra}, our method retains vivid large-scale clothing dynamics, fine clothing wrinkles and high-quality zoom-in levels of appearance details.
\par
\textbf{Geometry Synthesis.} Fig.~\ref{fig:gallray} presents the geometry synthesis results. 
Our approach generates an overall smooth surface, while preserving fine-scale, vivid wrinkles, which benefits downstream applications such as relighting, as shown in (Sec.~\ref{sec:system}).

\subsection{Comparison}
\noindent\textbf{Competing Methods.} We conduct comprehensive comparisons with previous animatable avatar methods, which broadly fall into two categories: \textbf{Server-based Approaches} and \textbf{Mobile-based Approaches}.
\paragraph{Server-based Approaches} Server-based approaches impose heavy computational and memory demands, limiting their deployment to server-grade hardware.
As shown in Tab.~\ref{tab:quantitive}, methods such as MeshAvatar~\cite{chen2024meshavatar}, GaussianAvatar~\cite{hu2023gaussianavatar}, Ani-Gaussians~\cite{li2024animatable}, ASH~\cite{Pang_2024_CVPR}, and~\cite{zhu2025ultra} remain impractical for resource-constrained devices.
Specifically, for the sake of fair comparison, we extend the resolution of the Gaussian Splat textures of ASH~\cite{Pang_2024_CVPR} from $256\times256$ to $768\times768$ to be aligned with the teacher model~\cite{zhu2025ultra}.

\paragraph{Mobile-based Approaches} In stark contrast to server-based approaches, mobile-based approaches, namely, 3DGS-Avatar~\cite{qian20233dgs} and TaoAvatar~\cite{chen2025taoavatar} significantly reduce computational and memory footprints ($\sim$100$\times$),
making real-time deployment feasible on consumer hardware, including personal computers equipped with mid-tier GPUs and VR glasses.
It is worth mentioning that the original implementation of TaoAvatar is mainly tailored to standing poses and struggles to train on long sequences with diverse motions. 
To improve its capacity for modeling a wider range of pose variations, we increase the number of Gaussian splat blendshapes to $128$. 
Moreover, we do not compare against SqueezeMe~\cite{iandola2025squeezeme}, as its official implementation is not publicly available.

\par \noindent\textbf{Quantitative Comparison.} 
Tab.~\ref{tab:quantitive} presents quantitative comparisons on novel-view and novel-pose synthesis against competing methods.
Our approach consistently outperforms most server-based solutions, particularly for the perceptual metrics.
Compared with state-of-the-art method UMA~\cite{zhu2025ultra}, we achieve comparable quantitative performance with over 1000$\times$ less computation and 10$\times$ fewer parameters.
%

%
%%%%%%%%%%%%%%%%%%%%%%%%%%%%%%%%%%%%%%
%
\begin{figure*}[tp]
    \centering
    \includegraphics[width=0.97\linewidth]{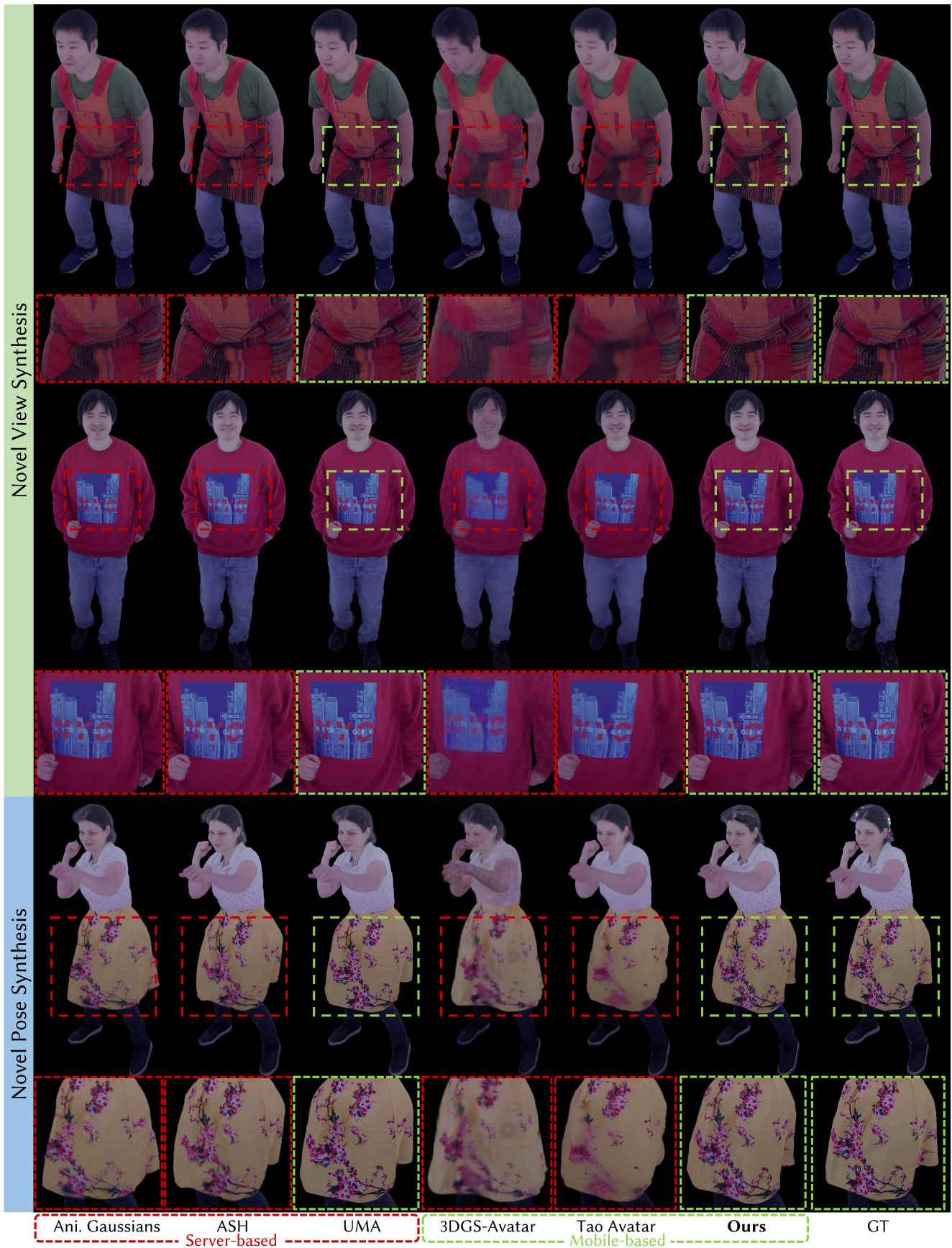}
    \vspace{-10pt}
    \caption{
        \textbf{Qualitative Comparison.} The qualitative comparison between our  method and the state-of-the-art approaches.
        Specifically, Animatable Gaussians, ASH and UMA belongs to server-based approaches, while 3DGS-Avatar and Tao Avatar belongs to mobile-based approaches.
        Please \textbf{zoom-in} to better inspect the detailed clothing wrinkles and appearances.
    }
    \label{fig:comparison}
\end{figure*}
%
%%%%%%%%%%%%%%%%%%%%%%%%%%%%%%%%%%%%%%
%
\begin{table*}[t]
\centering
\small
\caption{\textbf{Qualitative Comparison}. 
We quantitatively compare our method with the prior works on rendering accuracy on the training split and testing split across all sequences. 
We highlight the \fst{best}, \snd{second-best}, and \trd{third-best} scores.
}
\vspace{-3pt}
\begin{tabular}{|l|c|c|c|c|c|c|c|c|c|}
\hline
& \multicolumn{3}{c|}{Training Pose}
& \multicolumn{3}{c|}{Testing Pose}
& \multicolumn{1}{c|}{}
& \multicolumn{1}{c|}{} 
& \multicolumn{1}{c|}{} \\
\cline{2-7}
\multirow{-2}{*}{Methods} 
& \textbf{PSNR} $\uparrow$ 
& \textbf{SSIM} $\uparrow$ 
& \textbf{LPIPS} $\downarrow$ 
& \textbf{PSNR} $\uparrow$  
& \textbf{SSIM} $\uparrow$ 
& \textbf{LPIPS} $\downarrow$
& \multirow{-2}{*}{GFLOPS $\downarrow$} 
& \multirow{-2}{*}{Param (M) $\downarrow$}
& \multirow{-2}{*}{FPS $\uparrow$}
\\ 
\hline
MeshAvatar~\cite{chen2024meshavatar}       
& 27.23 & 0.8873 & 105.7 
& 25.98 & 0.8805 & 117.1 
& 518 & 39.51 & 1.1\\
GaussianAvatar~\cite{hu2023gaussianavatar}    
& 25.88 & 0.8884 & 127.2 
& 25.26 & 0.8845 & 134.9 
& 112.3 & \fsnd{5.77} & 34.6\\
Ani Gaussians~\cite{li2024animatable}        
& 29.07 & 0.9042 & 75.42 
& 26.06 & 0.8839 & \ftrd{103.3} 
& 2205.4 & 230.7 & 5.1 \\
ASH~\cite{Pang_2024_CVPR}       
& \fsnd{35.96} & \fsnd{0.9569} & \ftrd{63.84} 
& \ftrd{27.50} & \ftrd{0.8974} & 112.4 
& 1804 & 181.5 & 10.0 \\
\textbf{UMA}~\cite{zhu2025ultra}
& \ffst{36.80} & \ffst{0.9657} & \ffst{41.90} 
& \fsnd{27.66} & 0.8943 & \ffst{90.21} 
& 1804 & 184.0 & 9.9 \\
\hline
3DGS-Avatar~\cite{qian20233dgs}    
& 25.55 & 0.8865 & 141.6 
& 24.87 & 0.8822 & 146.7 
& \ftrd{6.84} & \ffst{3.23}  & \ftrd{35.2} \\
TaoAvatar~\cite{chen2025taoavatar}
& 28.66 & 0.9247 & 90.74 
& \ffst{28.01} & \ffst{0.9108} & 114.84 
& \fsnd{1.80} & 592  & \fsnd{87.8} \\
\textbf{Ours}       
& \ftrd{32.31} & \ftrd{0.9320} & \fsnd{53.88} 
& 27.31 & \fsnd{0.8978} & \fsnd{93.89} 
& \ffst{0.52} & \ftrd{26.67} & \ffst{182.2} \\
\hline
\end{tabular}
\label{tab:quantitive}
\end{table*}

Regarding mobile-based approaches such as TaoAvatar~\cite{chen2025taoavatar}, our method consistently achieves significantly better quantitative performance in both novel-view and novel-pose synthesis, while requiring fewer FLOPs and achieving faster runtime.
Notably, the TaoAvatar results reported here are based on the strengthened 128-blendshape setting introduced above.
Even under this stronger baseline, our method still delivers the best overall performance among mobile-based approaches.
This further demonstrates that the proposed \textit{Multi-level Spatial Factorized Blendshapes} effectively balances visual quality, computational efficiency, and memory consumption.

\par \noindent\textbf{Qualitative Comparison.}
Fig.~\ref{fig:comparison} shows qualitative comparisons for both novel-view and novel-pose rendering.
3DGS-Avatar~\cite{qian20233dgs} produces noticeably blurry renderings and struggle to recover motion-dependent wrinkle geometry and appearance details.
While Animatable Gaussians~\cite{li2024animatable} and ASH~\cite{Pang_2024_CVPR} capture motion-aware clothing wrinkles to some extent, they fail to reconstruct fine texture patterns.
UMA~\cite{zhu2025ultra} successfully recovers high-quality wrinkle geometry and detailed clothing patterns at the cost of extremely high computation overhead.
As a state-of-the-art mobile-friendly animatable avatar method, even with the extended number of blendshape bases, i.e., $128$, TaoAvatar~\cite{chen2025taoavatar} produces noticeably blurrier renderings, particularly for clothing undergoing large deformations, and also exhibits visible artifacts such as floating Gaussian splats.
In contrast, our method preserves high-fidelity geometric deformations and detailed wrinkles while requiring over 1000$\times$ fewer FLOPs than server-oriented methods and substantially fewer parameters, making it well suited for deployment on most mobile devices.

\subsection{Ablations}
\label{subsec:ablations}
 
\begin{table*}[t]
\normalsize
\centering
\caption{\textbf{Ablations}. 
We quantitatively compare our method with the ablative alternatives on rendering quality accuracy on the training split and testing split across all sequences. 
We highlight the \fst{best}, \snd{second-best}, and \trd{third-best} scores.}
\vspace{-3pt}
\begin{tabular}{|l|c|c|c|c|c|c|c|c|}
\hline
& \multicolumn{3}{c|}{Training Pose}
& \multicolumn{3}{c|}{Testing Pose}
& \multicolumn{1}{c|}{}
& \multicolumn{1}{c|}{} \\
\cline{2-7}
\multirow{-2}{*}{Methods} 
& \textbf{PSNR} $\uparrow$ 
& \textbf{SSIM} $\uparrow$ 
& \textbf{LPIPS} $\downarrow$ 
& \textbf{PSNR} $\uparrow$  
& \textbf{SSIM} $\uparrow$ 
& \textbf{LPIPS} $\downarrow$
& \multirow{-2}{*}{GFLOPS $\downarrow$} 
& \multirow{-2}{*}{Param (M) $\downarrow$}
\\ 
\hline
PCA-Only     & 29.04        & \fsnd{0.9308} & 68.19         & \ffst{24.67} & \ffst{0.8991} & 127.7         & \ftrd{1.93}   & 974.5 \\       
Single-Level Fact. & 29.10        & 0.9284        & 71.06         & \fsnd{24.19} & \fsnd{0.8952} & 127.8         & 4.43          & \ftrd{20.1} \\   
\hline
2D-Conv+2D-Conv & 29.16        & 0.9263        & 63.13         & 24.09        & 0.8887        & 118.8         & 13.7          & \ffst{18.22}  \\ 
2D-Conv+BS   & \ftrd{29.18}        & 0.9270        & \ftrd{62.61}         & 24.12        & \ftrd{0.8896}
& \fsnd{117.8}  & \fsnd{1.77}   & \fsnd{18.5} \\ 
\hline
BS+1D-Conv   & \fsnd{29.44} & \ftrd{0.9292} & \fsnd{61.22}  & 24.10        & 0.8889        & \ftrd{118.5}  & \ftrd{1.93}   & 26.52  \\       
\hline
\textbf{Ours} & \ffst{29.91} & \ffst{0.9330} & \ffst{58.43} & \ftrd{24.12} & 0.8894 & \ffst{117.7}  & \ffst{0.52}   & 26.7   \\ 
\hline
\end{tabular}
\label{tab:ablations}
\end{table*}

%
%%%%%%%%%%%%%%%%%%%%%%%%%%%%%%%%%%%%%%
%
\begin{figure*}[tp]
    \centering
    \includegraphics[width=\linewidth]{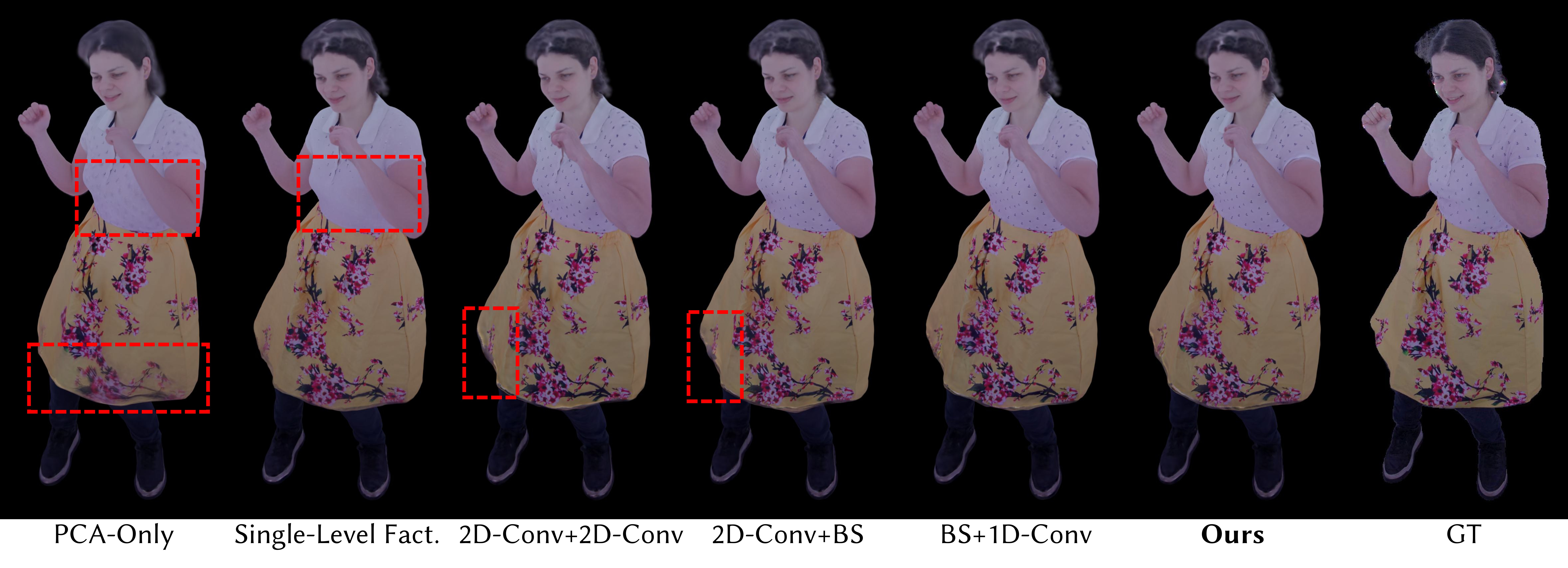}
    \vspace{-15pt}
    \caption{
    \textbf{Ablations}. Qualitative comparison of our full model against the design alternatives.
    Our full model better preserves wrinkle and appearance details with lowest computational overhead.
    }
    \label{fig:ablations}
\end{figure*}
%
%%%%%%%%%%%%%%%%%%%%%%%%%%%%%%%%%%%%%%
%

To assess the impact of our core design choices, we compare our full model against several design alternatives, where the ablations are organized around three questions:
(i) whether dynamic Gaussian splat textures should be modeled directly at full resolution without wavelet decomposition,
(ii) whether 2D convolutions are suitable for learning decomposed wavelet subbands,
and (iii) whether introducing localized motion-aware conditioning into 1D-factorized wavelet detail modeling can further improve performance.

We first investigate design alternatives that directly operate on the full-resolution Gaussian splat textures $\mathbf{T}^{\mathrm{gs}}_f$:

\noindent\textbf{PCA-Only.}
As discussed in Sec.~\ref{subsec:coarsegeodistall}, PCA~\cite{mackiewicz1993principal} is effective for compressing the coarse geometry of clothed humans~\cite{loper2015smpl,SMPL-X:2019}.
A straightforward extension is to apply it for the fine-scale geometry and appearance represented by the Gaussian splat textures $\mathbf{T}^{\mathrm{gs}}_f$.
Specifically, we construct a PCA subspace with 128 coefficients to compress the Gaussian splat textures $\mathbf{T}^{\mathrm{gs}}_f$ over all training frames, and train a lightweight MLP conditioned on the normalized skeletal motion $\bar{\boldsymbol{\theta}}_f$ to predict the corresponding coefficient weights.

\noindent\textbf{Single-level Factorization.}
To improve the scalability of learning blendshape-like correctives~\cite{chen2025taoavatar} over the high-resolution Gaussian texture $\mathbf{T}^{\mathrm{gs}}_f$, we adopt a single-stage factorized representation that models the dynamic Gaussian texture map as a weighted sum of outer products of learned 1D bases, which is introduced in Sec.~\ref{subsec:finegeodistill}.
For a stronger baseline, we set the numbers of blendshapes for geometry and appearance to $384$ and $256$, respectively.

As shown in Fig.~\ref{fig:ablations} and Tab.~\ref{tab:ablations}, both alternatives that directly model the full-resolution Gaussian splat textures exhibit clear limitations.
Modeling $\mathbf{T}^{\mathrm{gs}}_f$ with PCA results in a notably large number of parameters, and jointly representing detailed geometry and appearance in a single PCA subspace leads to blurry renderings and visible artifacts, particularly in regions with large clothing deformations.

Although the single-level factorized representation reduces the number of parameters even with a large number of blendshapes, it is still applied at full resolution, where one spatial dimension remains as large as $768$ and therefore still incurs considerable computational cost.
Moreover, jointly modeling low- and high-frequency components in a single stage limits the ability of the blendshape-like correctives to capture fine-scale appearance details, again resulting in blurry renderings.
These results suggest that directly modeling the full-resolution Gaussian splat textures is suboptimal, motivating us to instead represent the animatable clothed human using decomposed multi-level wavelet subbands.

We next move to the wavelet domain and investigate how the decomposed multi-level subbands should be learned:

\noindent\textbf{2D Conv (All Levels).}
As a straightforward baseline, we model all motion-dependent wavelet subbands as an image-to-image translation problem in texture space, following the teacher design in Sec.~\ref{subsec:teacher}.
Specifically, the LL subband $\mathbf{T}^{\mathrm{LL}}_{f}$ and the detail subbands $\{\mathbf{T}^{s}_{l,f}\}$ are predicted by separate 2D convolutional networks, each taking the positional and normal textures at the corresponding resolution $\{\mathbf{T}^{\mathrm{p}}_{l,f},\, \mathbf{T}^{\mathrm{n}}_{l,f}\}$:
\begin{align}
\mathbf{T}^{\mathrm{LL}}_{f} &= f^{\mathrm{LL}}_{\mathrm{conv}}\!\left(\mathbf{T}^{\mathrm{p}}_{3,f},\, \mathbf{T}^{\mathrm{n}}_{3,f}\right), \\
\left\{\mathbf{T}^{s}_{l,f}\right\}_{s \in \{\mathrm{LH}, \mathrm{HL}, \mathrm{HH}\}}
&= f^{l}_{\mathrm{conv}}\!\left(\mathbf{T}^{\mathrm{p}}_{l,f},\, \mathbf{T}^{\mathrm{n}}_{l,f}\right),
\end{align}
where $f^{\mathrm{LL}}_{\mathrm{conv}}(\cdot)$ and $f^{l}_{\mathrm{conv}}(\cdot)$ denote lightweight three-layer 2D convolutional networks with channel dimensions $32 \rightarrow 64 \rightarrow 128$, followed by a $1{\times}1$ convolution layer that decodes the Gaussian splat attributes from intermediate features.

\noindent\textbf{2D Conv (LL) + 1D Factorized Blendshapes (D2, D3).}
To further investigate whether the proposed blendshape formulation is more effective than 2D convolutions for the LL subband under a lightweight computational budget, we replace only the LL branch in our full model with the 2D convolutional network $f^{\mathrm{LL}}_{\mathrm{conv}}(\cdot)$ mentioned above.
The detail subbands $\mathbf{D}_2$ and $\mathbf{D}_3$ remain modeled by the original 1D factorized blendshape formulation, denoted as \textbf{2D-Conv + BS}.

As shown in Fig.~\ref{fig:ablations} and Tab.~\ref{tab:ablations}, applying 2D convolutions in the coarser wavelet domain is substantially cheaper than predicting Gaussian splat textures directly at the full resolution.
However, both variants still require approximately $20\times$ and $3\times$ more computation than our full model, respectively.
At the same time, they underperform our full model by a clear margin, particularly on perceptual metrics.

These results suggest that 2D convolutions are suboptimal for modeling wavelet subbands under a tight computational budget, which can be explained by the following reasons:
For the LL subband, the limited depth of the lightweight 2D convolutional network prevents it from establishing a sufficiently large receptive field to capture the global pose-dependent variations encoded in the low-frequency coefficients.
For the detail subbands, the inductive bias of 2D convolutions is poorly matched to their sparse, edge-aligned structure.

Finally, we investigate alternative strategies for learning the 1D-factorized wavelet detail subbands:

\noindent\textbf{2D Blendshapes (LL) + 1D Conv (D2, D3).}
Here, we keep the 2D blendshape formulation for the LL subband unchanged, while replacing the factorized blendshape representation for $\mathbf{D}_2$ and $\mathbf{D}_3$ with a spatially-aware 1D convolutional network, denoted as \textbf{BS+1D-Conv}.
Specifically, for each level $l \in \{2,3\}$, the position map $\mathbf{T}^{\mathrm{p}}_{f,l}$ is first projected into a feature space and concatenated channel-wise with sinusoidal positional encoding $\gamma(\mathbf{u})$.
The resulting features are then aggregated along the horizontal and vertical axes using learned pooling weights, producing compact 1D representations $\tilde{\mathbf{v}}_h$ and $\tilde{\mathbf{v}}_w$.
We then apply 1D convolutions along each axis to fuse local spatial information along each axis.
The refined 1D features along each axis are combined via an outer product and decoded by subband-specific $1{\times}1$ convolutions.

As seen in Fig.~\ref{fig:ablations} and Table~\ref{tab:ablations}, introducing additional local spatial awareness in the 1D factorized vectors does not improve over our full method.
This further proves that the learned blendshape bases already provide sufficient representation power for modeling the sparse detail subbands without the need for further spatially-aware feature extraction.

In contrast, our full method, denoted as \textbf{Ours}, is better aligned with the structural characteristics of different subbands:
All subbands are learned in a blendshape-driven manner to effectively capture pose-dependent variations, while the factorized outer-product representation enables compact and efficient modeling of the sparse, edge-aligned detail subbands.
As a result, our method achieves the best reconstruction quality with the lowest computational cost among all alternatives.

%
%%%%%%%%%%%%%%%%%%%%%%%%%%%%%%%%%%%%%%%%%%%%
%
\section{System} 
\label{sec:system}
%
%%%%%%%%%%%%%%%%%%%%%%%%%%%%%%%%%%%%%%
%
\begin{figure}[tp!]
    \centering
    \includegraphics[width=\linewidth]{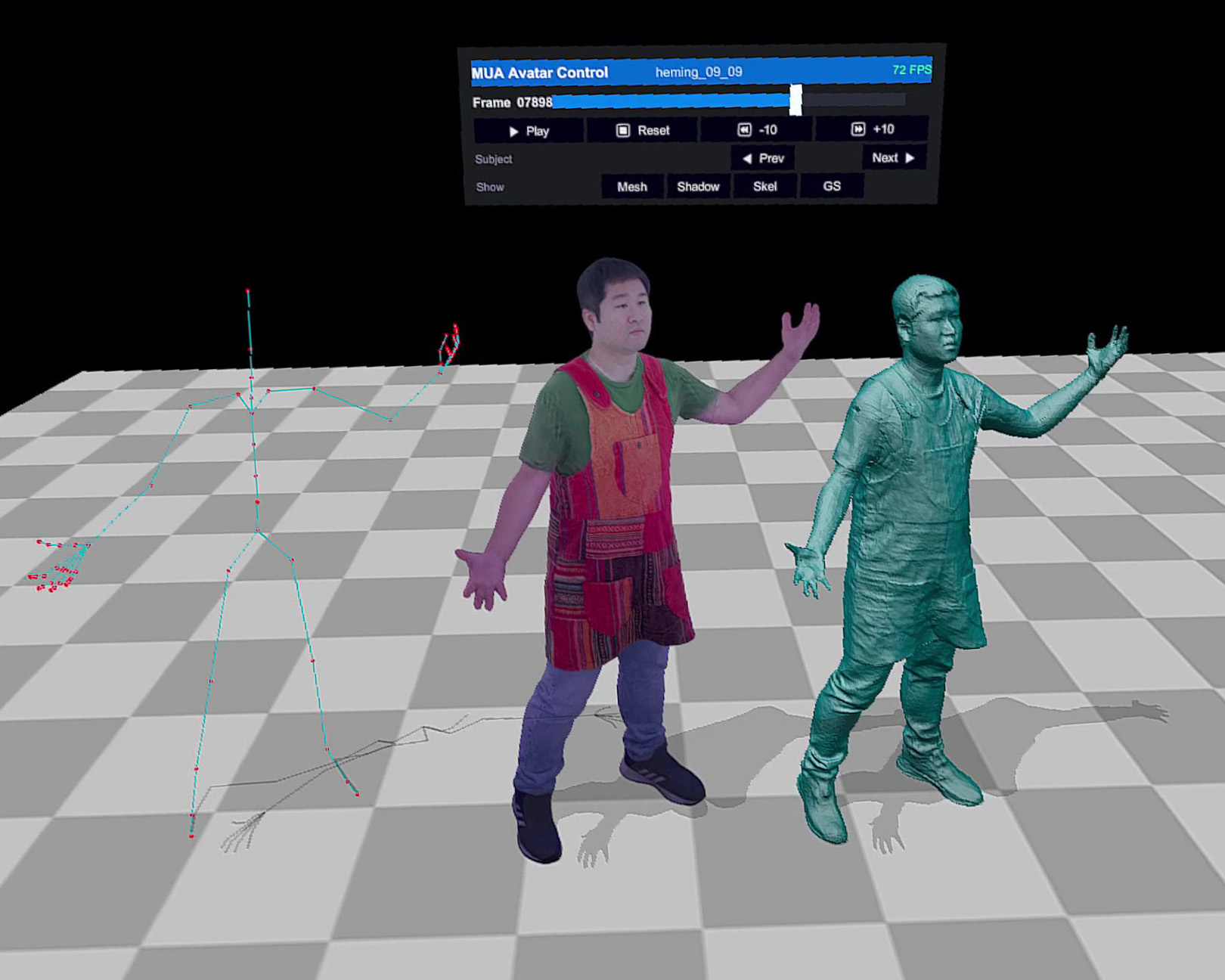}
    \vspace{-10pt}
    \caption{
    \textbf{Standalone VR demo.} Screenshot of our standalone VR demo running on Meta Quest 3.
    Users can inspect the dynamic avatar, detailed geometry, skeletal pose, and live shadows in the VR environment at a frame rate of $24$ FPS.
    All computations are performed on-the-fly on the headset.
    }
    \label{fig:system}
\end{figure}
%
%%%%%%%%%%%%%%%%%%%%%%%%%%%%%%%%%%%%%%
%
Fig.~\ref{fig:system} shows a screenshot of our standalone VR demo.
The system visualizes the dynamically rendered virtual avatar, along with its skeletal pose and detailed mesh geometry.
To further enhance the immersive experience, we additionally support rendering live shadows cast by the animated avatar.
Users can freely walk around the character in virtual space and use handheld controllers to interact with the system, including selecting frames for visualization, zooming in, toggling different visualizations, and switching between different characters.

\section{Limitations and Future Work}

Although MUA makes substantial progress towards ultra-detailed animatable human avatar modeling under limited computational and memory budgets, several challenges remain open.
First, since the teacher model (UMA) does not represent clothing as a separate layer, MUA currently does not support outfit swapping for the same subject.
A layered representation that disentangles garments from the human body could enable applications such as virtual try-on and wardrobe editing.
Second, as a learning-based approach, MUA does not explicitly model surface deformations induced by external physical forces, such as grasping interactions or wind effects.
Incorporating physics-based force models as additional geometric offsets could enhance realism and improve immersion.
Finally, the current model is driven solely by skeletal pose inputs.
Future work could explore multi-modal control signals, such as speech, text, or music rhythms, to enable richer and more expressive avatar animation.
\section{Conclusion}

We presented MUA, a mobile-friendly framework for bringing ultra-detailed animatable avatars to resource-constrained devices while maintaining the visual quality of a high-fidelity teacher model.
Our key idea is a wavelet-guided multi-level spatial factorized blendshape representation, which decomposes high-resolution Gaussian splat textures into multi-level frequency subbands and models each subband with a tailored formulation.
Based on this representation, we design a teacher-student distillation pipeline that substantially reduces both computational cost and model size, while preserving motion-dependent clothing dynamics and fine-grained appearance details.
Experiments demonstrate that MUA achieves a compelling quality-efficiency trade-off, clearly surpassing existing mobile-ready avatars and achieving comparable or superior quality to most server-oriented methods.
We further implement a complete system that achieves $90$ FPS on a desktop PC with streaming to Meta Quest 3, while also supporting fully standalone on-device execution at $24$ FPS.
We believe this represents a major step toward practical high-quality full-body clothed avatars that can run directly on consumer devices, and can facilitate future research on immersive digital humans for mobile and VR platforms.

\small
\bibliographystyle{IEEEtran}
\bibliography{IEEEabrv,main}

% Generated by IEEEtran.bst, version: 1.14 (2015/08/26)
\begin{thebibliography}{10}
\providecommand{\url}[1]{#1}
\csname url@samestyle\endcsname
\providecommand{\newblock}{\relax}
\providecommand{\bibinfo}[2]{#2}
\providecommand{\BIBentrySTDinterwordspacing}{\spaceskip=0pt\relax}
\providecommand{\BIBentryALTinterwordstretchfactor}{4}
\providecommand{\BIBentryALTinterwordspacing}{\spaceskip=\fontdimen2\font plus
\BIBentryALTinterwordstretchfactor\fontdimen3\font minus \fontdimen4\font\relax}
\providecommand{\BIBforeignlanguage}[2]{{%
\expandafter\ifx\csname l@#1\endcsname\relax
\typeout{** WARNING: IEEEtran.bst: No hyphenation pattern has been}%
\typeout{** loaded for the language `#1'. Using the pattern for}%
\typeout{** the default language instead.}%
\else
\language=\csname l@#1\endcsname
\fi
#2}}
\providecommand{\BIBdecl}{\relax}
\BIBdecl

\bibitem{chen2017exploring}
Z.~Chen, Y.~Wang, T.~Sun, X.~Gao, W.~Chen, Z.~Pan, H.~Qu, and Y.~Wu, ``Exploring the design space of immersive urban analytics,'' \emph{Visual Informatics}, vol.~1, no.~2, pp. 132--142, 2017.

\bibitem{sayffaerth2025educational}
C.~Sayffaerth, ``Educational twin: the influence of artificial xr expert duplicates on future learning,'' \emph{arXiv preprint arXiv:2504.13896}, 2025.

\bibitem{mildenhall2020nerf}
B.~Mildenhall, P.~P. Srinivasan, M.~Tancik, J.~T. Barron, R.~Ramamoorthi, and R.~Ng, ``Nerf: Representing scenes as neural radiance fields for view synthesis,'' in \emph{Eur. Conf. Comput. Vis.}, 2020.

\bibitem{kerbl20233d}
B.~Kerbl, G.~Kopanas, T.~Leimk{\"u}hler, and G.~Drettakis, ``3d gaussian splatting for real-time radiance field rendering,'' \emph{ACM Trans. Graph.}, vol.~42, no.~4, pp. 1--14, 2023.

\bibitem{neus2}
Y.~Wang, Q.~Han, M.~Habermann, K.~Daniilidis, C.~Theobalt, and L.~Liu, ``Neus2: Fast learning of neural implicit surfaces for multi-view reconstruction,'' in \emph{Int. Conf. Comput. Vis.}, 2023.

\bibitem{li2024animatable}
Z.~Li, Z.~Zheng, L.~Wang, and Y.~Liu, ``Animatable gaussians: Learning pose-dependent gaussian maps for high-fidelity human avatar modeling,'' in \emph{Proceedings of the IEEE/CVF conference on computer vision and pattern recognition}, 2024, pp. 19\,711--19\,722.

\bibitem{habermann2023hdhumans}
M.~Habermann, L.~Liu, W.~Xu, G.~Pons-Moll, M.~Zollhoefer, and C.~Theobalt, ``Hdhumans: A hybrid approach for high-fidelity digital humans,'' \emph{Proceedings of the ACM on Computer Graphics and Interactive Techniques}, vol.~6, no.~3, pp. 1--23, 2023.

\bibitem{li2022tava}
R.~Li, J.~Tanke, M.~Vo, M.~Zollhofer, J.~Gall, A.~Kanazawa, and C.~Lassner, ``Tava: Template-free animatable volumetric actors,'' 2022.

\bibitem{ARAH}
S.~Wang, K.~Schwarz, A.~Geiger, and S.~Tang, ``Arah: Animatable volume rendering of articulated human sdfs,'' in \emph{Eur. Conf. Comput. Vis.}, 2022.

\bibitem{Pang_2024_CVPR}
H.~Pang, H.~Zhu, A.~Kortylewski, C.~Theobalt, and M.~Habermann, ``Ash: Animatable gaussian splats for efficient and photoreal human rendering,'' in \emph{IEEE Conf. Comput. Vis. Pattern Recog.}, June 2024, pp. 1165--1175.

\bibitem{zhu2025ultra}
H.~Zhu, G.~Sun, C.~Theobalt, and M.~Habermann, ``Uma: Ultra-detailed human avatars via multi-level surface alignment,'' \emph{arXiv preprint arXiv:2506.01802}, 2025.

\bibitem{karaev2024cotracker}
N.~Karaev, I.~Rocco, B.~Graham, N.~Neverova, A.~Vedaldi, and C.~Rupprecht, ``Cotracker: It is better to track together,'' in \emph{European Conference on Computer Vision}.\hskip 1em plus 0.5em minus 0.4em\relax Springer, 2024, pp. 18--35.

\bibitem{iandola2025squeezeme}
F.~Iandola, S.~Pidhorskyi, I.~Santesteban, D.~Gupta, A.~Pahuja, N.~Bartolovic, F.~Yu, E.~Garbin, T.~Simon, and S.~Saito, ``Squeezeme: Mobile-ready distillation of gaussian full-body avatars,'' in \emph{Proceedings of the Special Interest Group on Computer Graphics and Interactive Techniques Conference Conference Papers}, 2025, pp. 1--11.

\bibitem{chen2025taoavatar}
J.~Chen, J.~Hu, G.~Wang, Z.~Jiang, T.~Zhou, Z.~Chen, and C.~Lv, ``Taoavatar: Real-time lifelike full-body talking avatars for augmented reality via 3d gaussian splatting,'' in \emph{Proceedings of the Computer Vision and Pattern Recognition Conference}, 2025, pp. 10\,723--10\,734.

\bibitem{SMPL-X:2019}
G.~Pavlakos, V.~Choutas, N.~Ghorbani, T.~Bolkart, A.~A.~A. Osman, D.~Tzionas, and M.~J. Black, ``Expressive body capture: 3d hands, face, and body from a single image,'' in \emph{IEEE Conf. Comput. Vis. Pattern Recog.}, 2019, pp. 10\,975--10\,985.

\bibitem{chen2022tensorf}
A.~Chen, Z.~Xu, A.~Geiger, J.~Yu, and H.~Su, ``Tensorf: Tensorial radiance fields,'' in \emph{European conference on computer vision}.\hskip 1em plus 0.5em minus 0.4em\relax Springer, 2022, pp. 333--350.

\bibitem{peng2021neuralbody}
S.~Peng, Y.~Zhang, Y.~Xu, Q.~Wang, Q.~Shuai, H.~Bao, and X.~Zhou, ``Neural body: Implicit neural representations with structured latent codes for novel view synthesis of dynamic humans,'' in \emph{IEEE Conf. Comput. Vis. Pattern Recog.}, 2021, pp. 9054--9063.

\bibitem{Lombardi2021MVP}
S.~Lombardi, T.~Simon, G.~Schwartz, M.~Zollhofer, Y.~Sheikh, and J.~M. Saragih, ``Mixture of volumetric primitives for efficient neural rendering,'' \emph{ACM Trans. Graph.}, vol.~40, no.~4, pp. 59:1--59:13, 2021.

\bibitem{wang2020learning}
Z.~Wang, T.~Bagautdinov, S.~Lombardi, T.~Simon, J.~Saragih, J.~Hodgins, and M.~Zollhofer, ``Learning compositional radiance fields of dynamic human heads,'' 2020.

\bibitem{weng_humannerf_2022_cvpr}
C.-Y. Weng, B.~Curless, P.~P. Srinivasan, J.~T. Barron, and I.~Kemelmacher-Shlizerman, ``Human{N}e{RF}: Free-viewpoint rendering of moving people from monocular video,'' in \emph{IEEE Conf. Comput. Vis. Pattern Recog.}, June 2022, pp. 16\,210--16\,220.

\bibitem{isik2023humanrf}
M.~I\c{s}{\i}k, M.~Runz, M.~Georgopoulos, T.~Khakhulin, J.~Starck, L.~Agapito, and M.~Niessner, ``Humanrf: High-fidelity neural radiance fields for humans in motion,'' \emph{ACM Trans. Graph.}, vol.~42, no.~4, pp. 1--12, 2023.

\bibitem{xu2024representing}
Z.~Xu, Y.~Xu, Z.~Yu, S.~Peng, J.~Sun, H.~Bao, and X.~Zhou, ``Representing long volumetric video with temporal gaussian hierarchy,'' \emph{ACM Transactions on Graphics (TOG)}, vol.~43, no.~6, pp. 1--18, 2024.

\bibitem{jiang2025reperformer}
Y.~Jiang, Z.~Shen, C.~Guo, Y.~Hong, Z.~Su, Y.~Zhang, M.~Habermann, and L.~Xu, ``Reperformer: Immersive human-centric volumetric videos from playback to photoreal reperformance,'' \emph{arXiv preprint arXiv:2503.12242}, 2025.

\bibitem{xiang2021modeling}
D.~Xiang, F.~Prada, T.~Bagautdinov, W.~Xu, Y.~Dong, H.~Wen, J.~Hodgins, and C.~Wu, ``Modeling clothing as a separate layer for an animatable human avatar,'' \emph{ACM Trans. Graph.}, vol.~40, no.~6, pp. 1--15, 2021.

\bibitem{alldieck18b}
T.~Alldieck, M.~Magnor, W.~Xu, C.~Theobalt, and G.~Pons-Moll, ``Detailed human avatars from monocular video,'' in \emph{International Conference on 3D Vision}, Sep. 2018, pp. 98--109.

\bibitem{alldieck19}
T.~Alldieck, M.~Magnor, B.~L. Bhatnagar, C.~Theobalt, and G.~Pons-Moll, ``Learning to reconstruct people in clothing from a single {RGB} camera,'' in \emph{IEEE Conf. Comput. Vis. Pattern Recog.}, Jun. 2019, pp. 1175--1186.

\bibitem{habermann2020deepcap}
M.~Habermann, W.~Xu, M.~Zollhofer, G.~Pons-Moll, and C.~Theobalt, ``Deepcap: Monocular human performance capture using weak supervision,'' in \emph{IEEE Conf. Comput. Vis. Pattern Recog.}, 2020, pp. 5052--5063.

\bibitem{habermann2019livecap}
M.~Habermann, W.~Xu, M.~Zollhoefer, G.~Pons-Moll, and C.~Theobalt, ``Livecap: Real-time human performance capture from monocular video,'' \emph{ACM Transactions On Graphics (TOG)}, vol.~38, no.~2, pp. 1--17, 2019.

\bibitem{xiu2023econ}
Y.~Xiu, J.~Yang, X.~Cao, D.~Tzionas, and M.~J. Black, ``{ECON: Explicit Clothed humans Optimized via Normal integration},'' in \emph{IEEE Conf. Comput. Vis. Pattern Recog.}, June 2023.

\bibitem{zhang2024sifu}
Z.~Zhang, Z.~Yang, and Y.~Yang, ``Sifu: Side-view conditioned implicit function for real-world usable clothed human reconstruction,'' in \emph{IEEE Conf. Comput. Vis. Pattern Recog.}, 2024, pp. 9936--9947.

\bibitem{zheng2025gstar}
C.~Zheng, L.~Xue, J.~Zarate, and J.~Song, ``Gstar: Gaussian surface tracking and reconstruction,'' \emph{arXiv preprint arXiv:2501.10283}, 2025.

\bibitem{zhu2022registering}
H.~Zhu, L.~Qiu, Y.~Qiu, and X.~Han, ``Registering explicit to implicit: Towards high-fidelity garment mesh reconstruction from single images,'' in \emph{IEEE Conf. Comput. Vis. Pattern Recog.}, 2022, pp. 3845--3854.

\bibitem{kwon2021neural}
Y.~Kwon, D.~Kim, D.~Ceylan, and H.~Fuchs, ``Neural human performer: Learning generalizable radiance fields for human performance rendering,'' \emph{Adv. Neural Inform. Process. Syst.}, 2021.

\bibitem{wang2021ibrnet}
Q.~Wang, Z.~Wang, K.~Genova, P.~Srinivasan, H.~Zhou, J.~T. Barron, R.~Martin-Brualla, N.~Snavely, and T.~Funkhouser, ``Ibrnet: Learning multi-view image-based rendering,'' in \emph{IEEE Conf. Comput. Vis. Pattern Recog.}, 2021.

\bibitem{Remelli2022TexelAligned}
E.~Remelli, T.~M. Bagautdinov, S.~Saito, C.~Wu, T.~Simon, S.~Wei, K.~Guo, Z.~Cao, F.~Prada, J.~M. Saragih, and Y.~Sheikh, ``Drivable volumetric avatars using texel-aligned features,'' in \emph{SIGGRAPH (Conference Paper Track)}, 2022, pp. 56:1--56:9.

\bibitem{shetty2023holoported}
A.~Shetty, M.~Habermann, G.~Sun, D.~Luvizon, V.~Golyanik, and C.~Theobalt, ``Holoported characters: Real-time free-viewpoint rendering of humans from sparse rgb cameras,'' in \emph{Proceedings of the IEEE/CVF Conference on Computer Vision and Pattern Recognition}, 2024, pp. 1206--1215.

\bibitem{sun2024metacap}
G.~Sun, R.~Dabral, P.~Fua, C.~Theobalt, and M.~Habermann, ``Metacap: Meta-learning priors from multi-view imagery for sparse-view human performance capture and rendering,'' in \emph{ECCV}, 2024.

\bibitem{sun2025real}
G.~Sun, R.~Dabral, H.~Zhu, P.~Fua, C.~Theobalt, and M.~Habermann, ``Real-time free-view human rendering from sparse-view rgb videos using double unprojected textures,'' June 2025.

\bibitem{zubekhin2025giga}
A.~Zubekhin, H.~Zhu, P.~Gotardo, T.~Beeler, M.~Habermann, and C.~Theobalt, ``Giga: Generalizable sparse image-driven gaussian humans,'' \emph{arXiv}, 2025.

\bibitem{blender}
\BIBentryALTinterwordspacing
{Blender Foundation}, ``Blender,'' 2025. [Online]. Available: \url{https://www.blender.org}
\BIBentrySTDinterwordspacing

\bibitem{stoll2010video}
C.~Stoll, J.~Gall, E.~De~Aguiar, S.~Thrun, and C.~Theobalt, ``Video-based reconstruction of animatable human characters,'' \emph{TOG}, vol.~29, no.~6, pp. 1--10, 2010.

\bibitem{guan2012drape}
P.~Guan, L.~Reiss, D.~A. Hirshberg, A.~Weiss, and M.~J. Black, ``Drape: Dressing any person,'' \emph{TOG}, vol.~31, no.~4, pp. 1--10, 2012.

\bibitem{xu2011video}
F.~Xu, Y.~Liu, C.~Stoll, J.~Tompkin, G.~Bharaj, Q.~Dai, H.-P. Seidel, J.~Kautz, and C.~Theobalt, ``Video-based characters: creating new human performances from a multi-view video database,'' in \emph{ACM SIGGRAPH 2011 papers}, 2011, pp. 1--10.

\bibitem{casas14}
D.~Casas, M.~Volino, J.~Collomosse, and A.~Hilton, ``4d video textures for interactive character appearance,'' \emph{Comput. Graph. Forum}, vol.~33, no.~2, p. 371–380, May 2014.

\bibitem{shysheya2019textured}
A.~Shysheya, E.~Zakharov, K.-A. Aliev, R.~Bashirov, E.~Burkov, K.~Iskakov, A.~Ivakhnenko, Y.~Malkov, I.~Pasechnik, D.~Ulyanov \emph{et~al.}, ``Textured neural avatars,'' in \emph{IEEE Conf. Comput. Vis. Pattern Recog.}, 2019, pp. 2387--2397.

\bibitem{bagautdinov2021driving}
T.~Bagautdinov, C.~Wu, T.~Simon, F.~Prada, T.~Shiratori, S.-E. Wei, W.~Xu, Y.~Sheikh, and J.~Saragih, ``Driving-signal aware full-body avatars,'' \emph{ACM Transactions on Graphics (TOG)}, vol.~40, no.~4, pp. 1--17, 2021.

\bibitem{xiang2022dressing}
D.~Xiang, T.~Bagautdinov, T.~Stuyck, F.~Prada, J.~Romero, W.~Xu, S.~Saito, J.~Guo, B.~Smith, T.~Shiratori \emph{et~al.}, ``Dressing avatars: Deep photorealistic appearance for physically simulated clothing,'' \emph{ACM Trans. Graph.}, vol.~41, no.~6, pp. 1--15, 2022.

\bibitem{habermann2021}
M.~Habermann, L.~Liu, W.~Xu, M.~Zollhoefer, G.~Pons-Moll, and C.~Theobalt, ``Real-time deep dynamic characters,'' \emph{ACM Trans. Graph.}, vol.~40, no.~4, aug 2021.

\bibitem{embedded}
R.~W. Sumner, J.~Schmid, and M.~Pauly, ``Embedded deformation for shape manipulation,'' \emph{ACM Trans. Graph.}, vol.~26, no.~3, p. 80–es, jul 2007.

\bibitem{chen2024meshavatar}
Y.~Chen, Z.~Zheng, Z.~Li, C.~Xu, and Y.~Liu, ``Meshavatar: Learning high-quality triangular human avatars from multi-view videos,'' in \emph{Eur. Conf. Comput. Vis.}\hskip 1em plus 0.5em minus 0.4em\relax Springer, 2024, pp. 250--269.

\bibitem{wang2021neus}
P.~Wang, L.~Liu, Y.~Liu, C.~Theobalt, T.~Komura, and W.~Wang, ``Neus: learning neural implicit surfaces by volume rendering for multi-view reconstruction,'' in \emph{Proceedings of the 35th International Conference on Neural Information Processing Systems}, 2021, pp. 27\,171--27\,183.

\bibitem{loper15}
M.~Loper, N.~Mahmood, J.~Romero, G.~Pons-Moll, and M.~J. Black, ``{SMPL}: A skinned multi-person linear model,'' \emph{ACM Trans. Graphics (Proc. SIGGRAPH Asia)}, vol.~34, no.~6, pp. 248:1--248:16, Oct 2015.

\bibitem{STAR:2020}
A.~Osman, T.~Bolkart, and M.~J. Black, ``Star: Sparse trained articulated human body regressor,'' in \emph{Eur. Conf. Comput. Vis.}, 2020, pp. 598--613.

\bibitem{TotalCapture2018}
H.~Joo, T.~Simon, and Y.~Sheikh, ``Total capture: A 3d deformation model for tracking faces, hands, and bodies,'' in \emph{IEEE Conf. Comput. Vis. Pattern Recog.}, 2018, pp. 8320--8329.

\bibitem{liu2021neural}
L.~Liu, M.~Habermann, V.~Rudnev, K.~Sarkar, J.~Gu, and C.~Theobalt, ``Neural actor: Neural free-view synthesis of human actors with pose control,'' \emph{ACM Trans. Graph.(ACM SIGGRAPH Asia)}, 2021.

\bibitem{peng2021animatable}
S.~Peng, J.~Dong, Q.~Wang, S.~Zhang, Q.~Shuai, X.~Zhou, and H.~Bao, ``Animatable neural radiance fields for modeling dynamic human bodies,'' in \emph{Int. Conf. Comput. Vis.}, 2021, pp. 14\,314--14\,323.

\bibitem{xu2021hnerf}
H.~Xu, T.~Alldieck, and C.~Sminchisescu, ``H-nerf: Neural radiance fields for rendering and temporal reconstruction of humans in motion,'' \emph{Adv. Neural Inform. Process. Syst.}, vol.~34, pp. 14\,955--14\,966, 2021.

\bibitem{NNA}
Q.~Gao, Y.~Wang, L.~Liu, L.~Liu, C.~Theobalt, and B.~Chen, ``Neural novel actor: Learning a generalized animatable neural representation for human actors,'' \emph{IEEE Trans. Vis. Comput. Graph.}, 2023.

\bibitem{zheng2023avatarrex}
Z.~Zheng, X.~Zhao, H.~Zhang, B.~Liu, and Y.~Liu, ``Avatarrex: Real-time expressive full-body avatars,'' \emph{ACM Trans. Graph.}, vol.~42, no.~4, 2023.

\bibitem{kwon2023deliffas}
Y.~Kwon, L.~Liu, H.~Fuchs, M.~Habermann, and C.~Theobalt, ``Deliffas: Deformable light fields for fast avatar synthesis,'' \emph{Adv. Neural Inform. Process. Syst.}, 2023.

\bibitem{zhu2023trihuman}
H.~Zhu, F.~Zhan, C.~Theobalt, and M.~Habermann, ``Trihuman: A real-time and controllable tri-plane representation for detailed human geometry and appearance synthesis,'' \emph{arXiv preprint arXiv:2312.05161}, 2023.

\bibitem{Chan2022}
E.~R. Chan, C.~Z. Lin, M.~A. Chan, K.~Nagano, B.~Pan, S.~D. Mello, O.~Gallo, L.~Guibas, J.~Tremblay, S.~Khamis, T.~Karras, and G.~Wetzstein, ``Efficient geometry-aware {3D} generative adversarial networks,'' in \emph{CVPR}, 2022.

\bibitem{ma2021scale}
Q.~Ma, S.~Saito, J.~Yang, S.~Tang, and M.~J. Black, ``Scale: Modeling clothed humans with a surface codec of articulated local elements,'' in \emph{CVPR}, 2021, pp. 16\,082--16\,093.

\bibitem{ma2021power}
Q.~Ma, J.~Yang, S.~Tang, and M.~J. Black, ``The power of points for modeling humans in clothing,'' in \emph{ICCV}, 2021, pp. 10\,974--10\,984.

\bibitem{lin2022learning}
S.~Lin, H.~Zhang, Z.~Zheng, R.~Shao, and Y.~Liu, ``Learning implicit templates for point-based clothed human modeling,'' in \emph{ECCV}.\hskip 1em plus 0.5em minus 0.4em\relax Springer, 2022, pp. 210--228.

\bibitem{loper2015smpl}
M.~Loper, N.~Mahmood, J.~Romero, G.~Pons-Moll, and M.~J. Black, ``Smpl: A skinned multi-person linear model,'' \emph{ACM Transactions on Graphics}, vol.~34, no.~6, 2015.

\bibitem{lei2023gart}
J.~Lei, Y.~Wang, G.~Pavlakos, L.~Liu, and K.~Daniilidis, ``Gart: Gaussian articulated template models,'' in \emph{CVPR}, 2024.

\bibitem{qian20233dgs}
Z.~Qian, S.~Wang, M.~Mihajlovic, A.~Geiger, and S.~Tang, ``3dgs-avatar: Animatable avatars via deformable 3d gaussian splatting,'' in \emph{CVPR}, 2024.

\bibitem{hu2023gauhuman}
S.~Hu and Z.~Liu, ``Gauhuman: Articulated gaussian splatting from monocular human videos,'' in \emph{CVPR}, 2024.

\bibitem{kocabas2023hugs}
M.~Kocabas, J.-H.~R. Chang, J.~Gabriel, O.~Tuzel, and A.~Ranjan, ``Hugs: Human gaussian splats,'' in \emph{CVPR}, 2024.

\bibitem{hu2023gaussianavatar}
L.~Hu, H.~Zhang, Y.~Zhang, B.~Zhou, B.~Liu, S.~Zhang, and L.~Nie, ``Gaussianavatar: Towards realistic human avatar modeling from a single video via animatable 3d gaussians,'' in \emph{CVPR}, 2024.

\bibitem{Ma_2021_CVPR}
S.~Ma, T.~Simon, J.~Saragih, D.~Wang, Y.~Li, F.~De~la Torre, and Y.~Sheikh, ``Pixel codec avatars,'' in \emph{CVPR}, June 2021, pp. 64--73.

\bibitem{bashirov2024morf}
R.~Bashirov, A.~Larionov, E.~Ustinova, M.~Sidorenko, D.~Svitov, I.~Zakharkin, and V.~Lempitsky, ``Morf: Mobile realistic fullbody avatars from a monocular video,'' in \emph{CVPR}, 2024, pp. 3545--3555.

\bibitem{SplattingAvatar:CVPR2024}
Z.~Shao, Z.~Wang, Z.~Li, D.~Wang, X.~Lin, Y.~Zhang, M.~Fan, and Z.~Wang, ``{SplattingAvatar: Realistic Real-Time Human Avatars with Mesh-Embedded Gaussian Splatting},'' in \emph{Computer Vision and Pattern Recognition (CVPR)}, 2024.

\bibitem{captury}
{TheCaptury}, ``{The Captury},'' \url{http://www.thecaptury.com/}, 2020.

\bibitem{kavan2007skinning}
L.~Kavan, S.~Collins, J.~{\v{Z}}{\'a}ra, and C.~O'Sullivan, ``Skinning with dual quaternions,'' in \emph{Proceedings of the 2007 symposium on Interactive 3D graphics and games}, 2007, pp. 39--46.

\bibitem{ronneberger2015u}
O.~Ronneberger, P.~Fischer, and T.~Brox, ``U-net: Convolutional networks for biomedical image segmentation,'' in \emph{Medical Image Computing and Computer-Assisted Intervention--MICCAI 2015: 18th International Conference, Munich, Germany, Oct 5-9, 2015, Proceedings, Part III 18}.\hskip 1em plus 0.5em minus 0.4em\relax Springer, 2015, pp. 234--241.

\bibitem{wang2018image}
Y.~Wang, X.~Tao, X.~Qi, X.~Shen, and J.~Jia, ``Image inpainting via generative multi-column convolutional neural networks,'' \emph{Advances in neural information processing systems}, vol.~31, 2018.

\bibitem{mackiewicz1993principal}
A.~Ma{\'c}kiewicz and W.~Ratajczak, ``Principal components analysis (pca),'' \emph{Computers \& Geosciences}, vol.~19, no.~3, pp. 303--342, 1993.

\bibitem{hu2022lora}
E.~J. Hu, Y.~Shen, P.~Wallis, Z.~Allen-Zhu, Y.~Li, S.~Wang, L.~Wang, W.~Chen \emph{et~al.}, ``Lora: Low-rank adaptation of large language models.'' \emph{Iclr}, vol.~1, no.~2, p.~3, 2022.

\bibitem{paszke2017automatic}
A.~Paszke, S.~Gross, S.~Chintala, G.~Chanan, E.~Yang, Z.~DeVito, Z.~Lin, A.~Desmaison, L.~Antiga, and A.~Lerer, ``Automatic differentiation in pytorch,'' in \emph{NIPS-W}, 2017.

\bibitem{kingma2017adam}
D.~P. Kingma and J.~Ba, ``Adam: A method for stochastic optimization,'' 2017.

\bibitem{liang2024analyticsplatting}
Z.~Liang, Q.~Zhang, W.~Hu, Y.~Feng, L.~Zhu, and K.~Jia, ``Analytic-splatting: Anti-aliased 3d gaussian splatting via analytic integration,'' 2024.

\bibitem{kleinbeck_multi-layer_2025}
\BIBentryALTinterwordspacing
C.~Kleinbeck, H.~Schieber, K.~Engel, R.~Gutjahr, and D.~Roth, ``Multi-layer gaussian splatting for immersive anatomy visualization,'' pp. 1--11. [Online]. Available: \url{https://ieeexplore.ieee.org/document/10919012}
\BIBentrySTDinterwordspacing

\bibitem{khirodkar2024sapiens}
R.~Khirodkar, T.~Bagautdinov, J.~Martinez, S.~Zhaoen, A.~James, P.~Selednik, S.~Anderson, and S.~Saito, ``Sapiens: Foundation for human vision models,'' in \emph{European Conference on Computer Vision}.\hskip 1em plus 0.5em minus 0.4em\relax Springer, 2024, pp. 206--228.

\bibitem{zhang2018perceptual}
R.~Zhang, P.~Isola, A.~A. Efros, E.~Shechtman, and O.~Wang, ``The unreasonable effectiveness of deep features as a perceptual metric,'' in \emph{IEEE Conf. Comput. Vis. Pattern Recog.}, 2018.

\end{thebibliography}

\end{document}